\documentclass[review]{elsarticle}

\usepackage[utf8]{inputenc}
\usepackage{url}
\usepackage{adjustbox}
\usepackage{amsfonts}
\usepackage{subfig}
\usepackage{amsmath}
\usepackage{comment}
\usepackage[pdftex,dvipsnames]{xcolor}  

\newcommand{\method}{$DuPLO$}
\newcommand{\smallCNN}{$SCNN$}

\begin{document}

\begin{frontmatter}

\title{DuPLO: A DUal view Point deep Learning architecture for time series classificatiOn}

\author{Roberto Interdonato\corref{cor1}}
\ead{roberto.interdonato@cirad.fr}
\address{CIRAD, UMR TETIS, Montpellier}

\author{Dino Ienco}
\ead{dino.ienco@irstea.fr}
\address{IRSTEA, UMR TETIS\\LIRMM, University of Montpellier, Montpellier}

\author{Raffaele Gaetano}
\ead{raffaele.gaetano@cirad.fr}
\address{CIRAD, UMR TETIS, Montpellier}

\author{Kenji Ose}
\ead{kenji.ose@irstea.fr}
\address{IRSTEA, UMR TETIS, Montpellier}

\cortext[cor1]{Corresponding author}


\address{}

\begin{abstract}
Nowadays, modern Earth Observation systems continuously generate huge amounts of data.  A notable example is represented by the Sentinel-2 mission, which provides images at high spatial resolution (up to 10m) with high temporal revisit period (every 5 days), which can be organized in Satellite Image Time Series (SITS). While the use of SITS has been proved to be beneficial in the context of Land Use/Land Cover (LULC) map generation, unfortunately, machine learning approaches commonly leveraged in remote sensing field fail to take advantage of spatio-temporal dependencies present in such data.

Recently, new generation deep learning methods allowed to significantly advance research in this field. These approaches have generally focused on a single type of neural network, i.e., Convolutional Neural Networks (CNNs) or Recurrent Neural Networks (RNNs), which  model different but complementary information: spatial autocorrelation (CNNs) and temporal dependencies (RNNs). 
In this work, we propose the first deep learning architecture for the analysis of SITS data, namely \method{} (DUal view Point deep  Learning  architecture  for  time  series  classificatiOn),  that combines Convolutional and Recurrent neural networks to exploit their complementarity. Our hypothesis is that,  since CNNs and RNNs capture different aspects of the data, a combination of both models would produce a more diverse and complete representation of the information for the underlying land cover classification task.
Experiments carried out on two study sites characterized by different land cover characteristics  (i.e.,  the \textit{Gard} site  in  France  and  the \textit{Reunion  Island} in  the Indian Ocean), demonstrate the significance of our proposal.

\end{abstract}


\end{frontmatter}

\section{Introduction}
\label{sec:intro}

Modern Earth Observation (EO) systems produce huge volumes of data every day, involving programs that provide satellite images at high spatial resolution with high temporal revisit period.  High-resolution Satellite Image Time Series (SITS) represent a practical way to organize this information, which is particularly useful for area monitoring tasks.
A notable example is the Sentinel-2 mission, which is part of the Copernicus Programme, i.e.,  a programme developed by the European Space Agency (ESA) that involves a constellation of satellites monitoring different aspects of the Earth surface.
The Sentinel-2 mission allows to monitor the entire Earth Surface at 10m of spatial resolution with a revisit period between 5 and 10 days, supplying optical information ranging from visible to near and medium infrared. One of the main advantages of this mission is that the produced data are publicly available.

For these reasons, the use of SITS is gaining increasing success in a plethora of different domains, such as ecology, agriculture, mobility, health, risk monitoring and land management planning~\cite{BegueABBAFLLSV18,OlenB18,KoleckaGPPV18,ChenJMCYCX14,KhialiIT18,guttler2017,BellonBSAS17,IngladaVATMR17,KussulLSS17}. 
In the context of  Land Use/Land Cover (LULC) classification, exploiting SITS can be fruitful to discriminate among classes that exhibit different temporal behaviors~\cite{Abade15}, i.e., with the respect to the results that can be obtained using a single image. 
In~\cite{BellonBSAS17}, the authors propose to exploit SITS data to extract homogeneous land units in terms of phenological patterns and, later, for the automatic classification of land units according to their land-cover. 
The effectiveness of Sentinel-2 SITS to produce land cover maps at country scale has been showed in~\cite{IngladaVATMR17}, demonstrating  the practical interest of such data source. 
In~\cite{KussulLSS17}, the authors combine multi-source optical (Landsat-8) and radar (Sentinel-1) SITS in order to improve land cover maps on the agricultural domain. Another example is supplied in~\cite{KoleckaGPPV18} where optical SITS are leveraged to characterize grassland area as proxy indicator for biodiversity, food production, and global carbon cycle. 

Despite the usefulness of temporal trends that can be derived from remote sensing time series, most of the strategies proposed for SITS analysis tasks~\cite{Flamary15,Heine16,IngladaVATMR17,BellonBSAS17}, directly apply standard machine learning approaches (i.e. Random Forest, SVM) on the stacked images, thus ignoring any temporal dependencies that may be discovered in the data. Indeed, such algorithms make the assumption that the information (spectral bands and timestamps) are independent from each other.

Recently, the deep learning revolution~\cite{Zhang16} has shown that neural network models are well adapted tools for automatically managing and classifying remote sensing data.
The main characteristic of these models is the ability to simultaneously extract features optimized for image classification as well as the associated classifier. Moreover, unlike standard machine learning approaches, they can be used to discover spatial and temporal dependencies in SITS data.
Deep learning methods can be roughly categorized in two families of techniques: convolutional neural networks~\cite{Zhang16} (CNNs) and recurrent neural networks~\cite{BengioCV13} (RNNs).  
CNNs are well suited to model the spatial autocorrelation available in an image and they are already a well-known tool in the field of Remote Sensing~\cite{Zhang16,KussulLSS17,Zhu17}. Conversely, RNNs are specifically tailored to manage time dependencies~\cite{IencoGDM17} from multidimensional time series and they are recently starting to get attention in the Remote Sensing community~\cite{IencoGDM17,Lyu16,abs-1803-01945}. Such models explicitly capture temporal correlations by recursion and they have already proved to be effective in different domains such as speech recognition~\cite{Graves13}, natural language processing~\cite{LinzenDG16} and image completion~\cite{OordKEKVG16}.

Considering the analysis of SITS data, few works already exist which exploit deep learning to analyze such kind of data. A CNN based strategy is employed in~\cite{KussulLSS17} to deal with land cover classification in the agricultural domain. The main idea is to obtain a single image by stacking all the images in an input SITS, then using it to fed a CNN-based model. The results demonstrate the quality of the proposed approach w.r.t. a standard machine learning algorithm (i.e., Random Forest classifier).
In~\cite{Lyu16} a binary change detection classification task (i.e., change vs. no-change) has been addressed using a RNN model on a small time series of two dates.
The authors in~\cite{IencoGDM17} propose a RNN-based approach for land cover classification on optical SITS data. The work evaluates deep learning methods on very high resolution and high resolution data on two different study sites. These preliminary results have paved the way to the use of such models for the analysis of Satellite image time series data. Always in the context ofh land cover classification tasks,
RNN have also been used for the analysis of radar SITS data~\cite{MinhIGLNOM18,NdikumanaMBCH18}.

Even if we acknowledge the existence of a significant body of work dealing with the use of deep learning for the analysis of SITS data, at the same time it should be observed how previous works were tied to the choice of a specific network model, i.e., focusing either on Recurrent or Convolutional Neural Networks.
Even though both approaches have been shown to be effective on the analysis of SITS data, our hypothesis is that,  since they capture different knowledge aspects, a combination of both would produce a more diverse and complete representation of the information.
For this reason, we propose a deep learning architecture for the analysis of SITS data, namely \method{} (DUal view Point deep Learning architecture for time series classificatiOn), based on the combination of Convolutional and Recurrent neural networks. 
To the best of our knowledge, \method{} is the first example of deep learning architecture combining RNN and CNN approaches for the analysis of satellite image time series.

The idea behind \method{} is to take advantage of the fact that the two strategies (i.e., CNN and RNN) focus on different characteristics of the data, so that addressing the problem from a dual view point will allow to exploit as much as possible the complementary information produced by such models. 
Experiments carried out on two study sites characterized by different land cover characteristics (i.e., the \textit{Gard} site in France and the \textit{Reunion Island} in the Indian Ocean), demonstrate the effectiveness of our proposal compared to state of the art approaches for the characterization of land cover mapping on SITS data. Furthermore, the quantitative and qualitative results emphasize how the combination of CNN and RNN is beneficial for the classification task compared to the use of a single neural network model. 

The rest of the article is structured as follows: Section~\ref{sec:method} introduces the proposed deep learning architecture for land cover classification from SITS data. The study sites and the associated data are presented in Section~\ref{sec:data} while, the experimental setting and the evaluations are carried out and discussed in Section~\ref{sec:expe}. Finally, Section~\ref{sec:conclu} draws conclusions.

\section{\method{}: A DUal view Point deep Learning architecture for time series classificatiOn}
\label{sec:method}

\begin{figure}[t]
\centering
\includegraphics[width=\columnwidth]{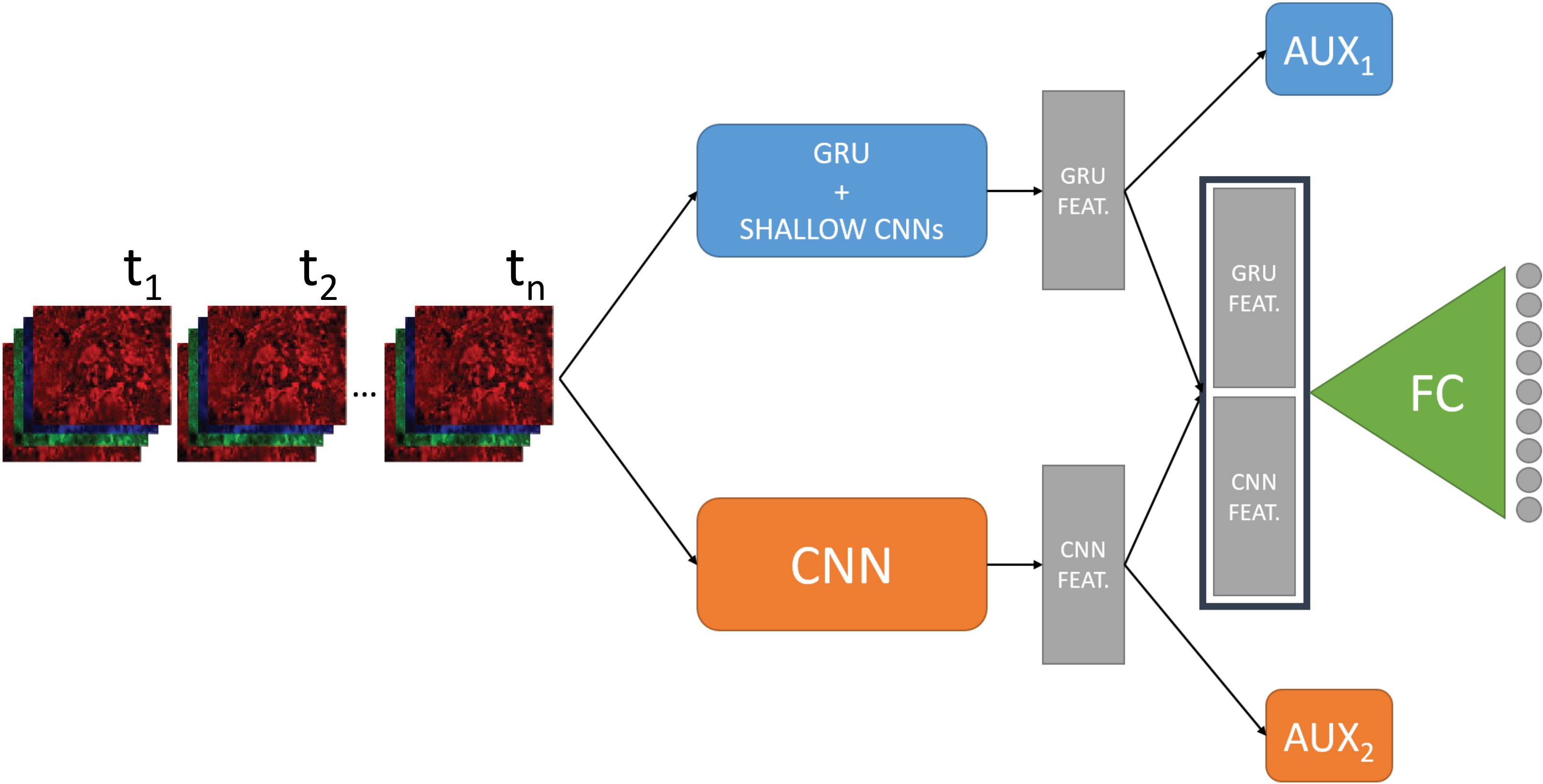}
\caption{ \label{fig:duplo} Visual representation of the  \method{} Deep Learning Architecture.}
\end{figure}

Figure~\ref{fig:duplo} shows a visual representation of the proposed \method{} deep learning architecture for the satellite image time series classification process. The optical time series of Sentinel-2 (S2) satellite images in input is first processed independently by the two branches consisting of two different neural networks: a CNN (cf. Section~\ref{sec:cnn}) and a RNN (cf. Section~\ref{sec:rnn}). Each branch supplies complementary information for the discriminative process since they look at the information from different points of view. The result of each branch is a feature vector summarizing the extracted knowledge. Each vector is used independently to train an auxiliary classifier (cf. Section~\ref{sec:train}), while the concatenation of both feature vectors (i.e., in a single feature vector of 2048 descriptors) is used to fed the final classifier that produces the land cover decision.

\subsection{CNN Branch}
\label{sec:cnn}

\begin{figure}[t]
\centering
\includegraphics[scale=.5]{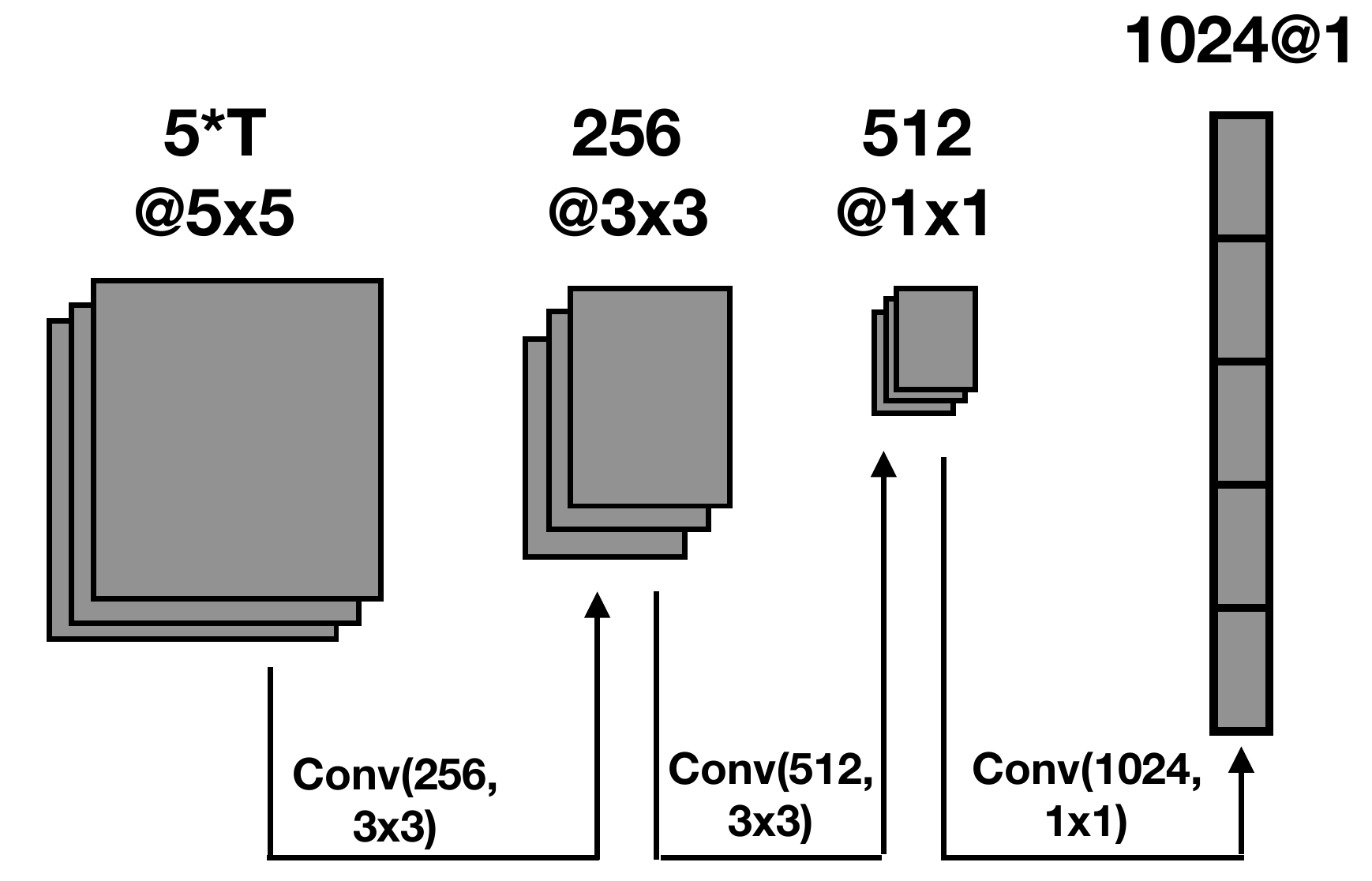}
\caption{ \label{fig:FCNN} Visual representation of the CNN module representing the branch of the \method{} model that manages the whole time series information as a stacked image. The $T$ value indicates the number of timestamps in the satellite image time series.}
\end{figure}

Figure~\ref{fig:FCNN} depicts the network architecture associated to the CNN branch of \method{}.
For this branch, 
we took inspiration from the VGG model~\cite{SimonyanZ14a}, a well-known network architecture usually adopted to tackle with standard Computer Vision tasks. The basic idea behind this model is to constantly increase the number of filters along the network, as long as a reasonable size of the feature maps has been reached. 

Inspecting this architecture, we can observe that the first convolution has a kernel of 3$\times$3 and it produces 256 feature maps. The second convolution still applies a kernel of size 3$\times$3 outputting 512 feature maps while the last convolution, with kernel size 1$\times$1, is only devoted to increase the final number of extracted features to 1024. All the convolutions are associated with a linear filter, followed by a Rectifier Linear Unit (ReLU) activation function~\cite{NairH10} to induce non-linearity and a batch normalization step~\cite{IoffeS15} that accelerates deep network training convergence by reducing the internal covariate shift.

The ReLU activation function is defined as follows:
\begin{align}
ReLU(x) &= Max(0, W \cdot x + b) \label{eqn:relu}
\end{align}
This activation function is defined on the positive part of the linear transformation of its argument ($W \cdot x + b$) where $x$ is the input information and $W$ and $b$ are parameters learned by the neural network model.  The choice of ReLU nonlinearities is motivated by two factors: i) the good convergence properties it guarantees and ii) the low computational complexity it provides~\cite{NairH10}.

Even though the proposed architecture shows a reasonable number of parameters, the training of such models may be difficult and the final model can suffer by overfitting~\cite{DahlSH13}. In order to avoid such phenomena, following a common practice for the training of deep learning architectures, we add Dropout~\cite{DahlSH13} after each batch normalization step. We set the drop rate equal to 0.4 meaning that 40\% of the neurons are randomly deactivated at each propagation step, at training time.


\subsection{RNN Branch}
\label{sec:rnn}

Recurrent Neural Networks are well established machine learning techniques that demonstrate their quality in different domains such as speech recognition, signal and natural language processing~\cite{SomaMSFN15,LinzenDG16}. Unlike standard feed forward networks (e.g., Convolutional Neural Networks -- CNNs), RNNs explicitly manage temporal data dependencies since the output of the neuron at time t-1 is used, together with the next input, to feed the neuron itself at time t. 

Recently, recurrent neural network (RNN) approaches have been successfully applied to tasks in 
the remote sensing field, e.g.,  to produce land use mappings from time series of optical images~\cite{IencoGDM17} and to recognize vegetation cover status using Sentinel-1 radar time series~\cite{Dinh18}. Motivated by these recent research results, we introduce a RNN module to manage information from the Sentinel-2 time series with the aim to extract an alternative representation from the data. In our model, we choose the GRU unit (Gated Recurrent Unit) introduced in~\cite{ChoMGBBSB14} since it has a moderate number of parameters to learn and its effectiveness in the field of remote sensing has already been proved~\cite{IencoGDM17,MouGZ17}. Due to the fact that we consider patches of satellite images, centered around a central pixel, we do not use the GRU unit directly on the radiometric information. First, we use a shallow CNN, we name \smallCNN, to process the patches at each timestamp and, subsequently, we feed the RNN model with the information extracted by the convolutional models.
Finally, we couple the Gated Recurrent Unit with an \textit{attention} mechanism~\cite{BritzGL17}. 

\begin{figure}[t]
\centering
\includegraphics[scale=.5]{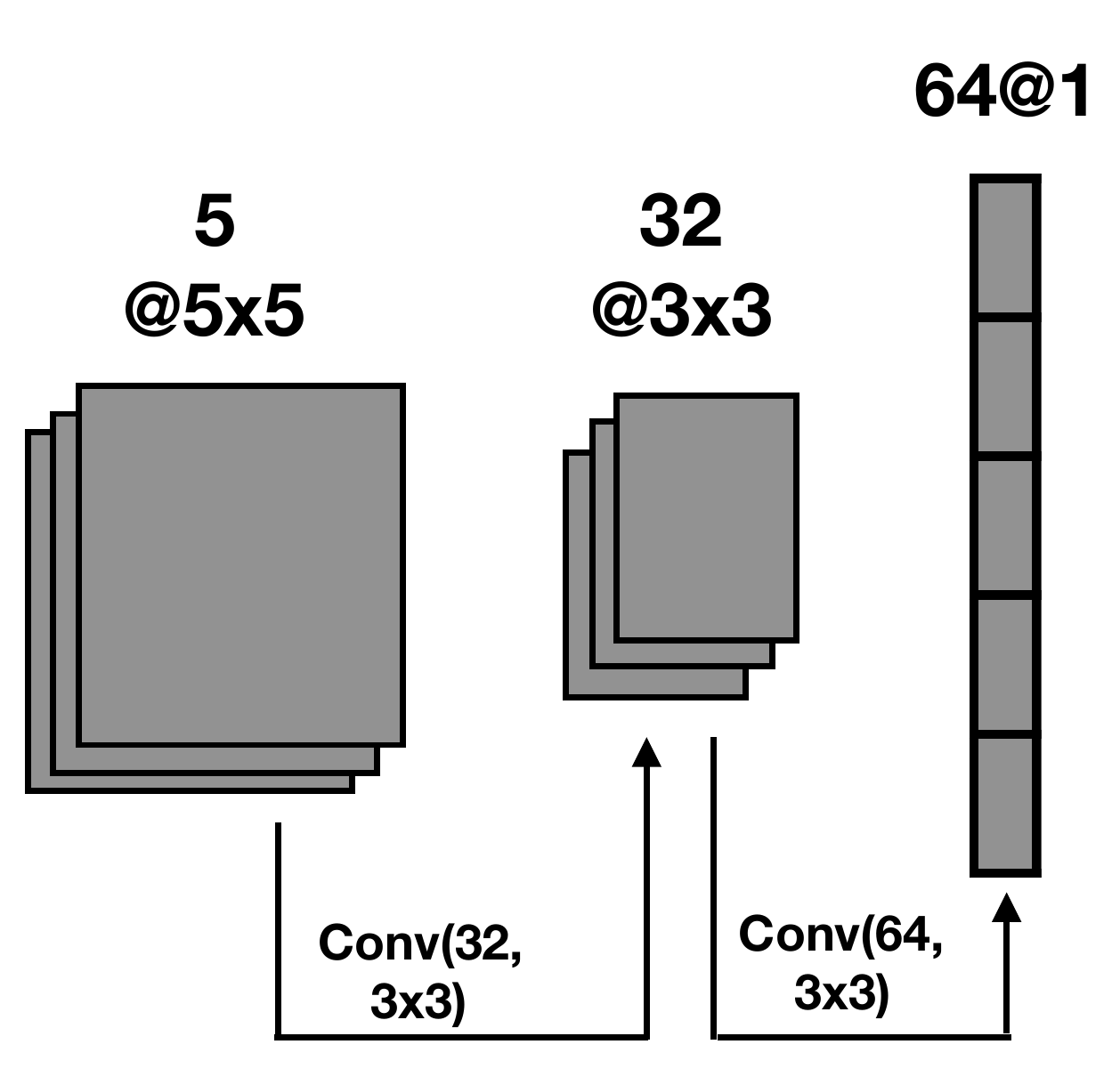}
\caption{ \label{fig:SCNN} Visual representation of the Shallow CNN applied at each timestamp of the satellite image time series. The results of this non-linear transformation is successively used to feed the Gate Recurrent Unit.}
\end{figure}

The structure of the \smallCNN is reported in Figure~\ref{fig:SCNN}. This shallow network is composed only by two convolutional layers, with 32 and 64 filters respectively, producing an output vector composed by 64 features. This step allows to extract the information carried out by the spatial neighborhood of the considered pixel before considering the temporal behavior of the satellite image time series. Also in this case, each convolution is associated with a linear filter, followed by a Rectifier Linear Unit (ReLU) activation function~\cite{NairH10} to induce non-linearity and a batch normalization step~\cite{IoffeS15}.

Once the \smallCNN is applied on the patches describing the time series, the input of a RNN unit is a sequence ($x_{t_1}$,..., $x_{t_T}$) where a generic element $x_{t_i}$ is a feature vector of cardinality equals to 64 (extracted by \smallCNN) and $t_i$ refers to the corresponding time stamp. 

Equations \ref{eqn:gru1}, \ref{eqn:gru2} and \ref{eqn:gru3} formally describe the \textit{GRU} neuron. 

\begin{align}
z_{t_i} &= \sigma(W_{zx} x_{t_i} + W_{zh} h_{t_{i-1}} + b_z  ) \label{eqn:gru1} \\
r_{t_i} &= \sigma(W_{rx} x_{t_i} + W_{rh} h_{t_{i-1}} + b_r  ) \label{eqn:gru2} \\
h_{t_i} &= z_t \odot h_{t-1} +  \label{eqn:gru3} \\
& (1-z_{t_i}) \odot \tanh( W_{hx} x_{t} + W_{hr} (r_t \odot h_{t_{i-1}})+b_h  ) \nonumber
\end{align}

The $\odot$ symbol indicates an element-wise multiplication while $\sigma$ and $\tanh$ represent Sigmoid and Hyperbolic Tangent function, respectively.

The \textit{GRU} unit has two gates, update ($z_t$) and reset ($r_t$), and one cell state, i.e., the hidden state ($h_t$). 
Moreover, the two gates combine the current input ($x_t$) with the information coming from the previous time stamp ($h_{t-1}$).
The update gate effectively controls the trade off between how much information from the previous hidden state will carry over to the current hidden state and how much information of the current time stamp needs to be kept.
On the other hand, the reset gate monitors how much information from the previous timestamps needs to be integrated with  current information. 
As all hidden units have separated reset and update gates, they are able to capture dependencies over different time scales. Units more prone to capture short-term dependencies will tend to have a frequently activated reset gate, but those that capture longer-term dependencies will have  update gates  that remain mostly active~\cite{ChoMGBBSB14}. This behavior enables the GRU unit to remember long-term information. 

Attention mechanisms~\cite{BritzGL17} are widely used in automatic signal processing (1D signal or language) and they allow to join together the information extracted by the GRU model at different time stamps. The output returned by the GRU model is a sequence of learned feature vectors for each time stamp: ($h_{t_1}$,..., $h_{t_N}$) where each $h_{t_i}$ has the same dimension $d$. Their matrix representation $H \in \mathbb{R}^{T,d}$ is obtained  vertically stacking the set of vectors.
The attention mechanism allows us to combine together these different vectors $h_{t_{i}}$, in a single one $rnn_{feat}$, to attentively combine the information returned by the GRU unit at each of the different time stamps. The attention formulation we use, starting from a sequence of vectors encoding the learned descriptors ($h_{t_1}$,..., $h_{t_T}$), is the following one:
\begin{align}
v_{a} &= tanh( H \cdot W_{a} + b_{a}) \label{eqn:att1} \\
\omega &= SoftMax(v_{a} \cdot u_{a}) \label{eqn:att2} \\
rnn_{feat} &= \sum_{i=1}^{T} \lambda_i \cdot h_{t_{i}} \label{eqn:att3}
\end{align}
where matrix $W_{a} \in \mathbb{R}^{d,d}$ and vectors $b_{a}, u_{a} \in \mathbb{R}^{d}$ are parameters learned during the process. These parameters allow to combine the vectors contained in matrix $H$.
The purpose of this procedure is to learn a set of weights ($\omega_{t_1}$,..., $\omega_{t_T}$) that allows the contribution of each time stamp to be weighted by $h_{t_i}$ through a linear combination. The $SoftMax(\cdot)$~\cite{IencoGDM17} function is used to normalize weights $\omega$ so that their sum is equal to 1. The output of the RNN module is the feature vector $rnn_{feat}$: it encodes temporal information related to $ts_i$ for the pixel $i$.

With the aim to supply an equal amount of information from each branch of analysis of the \method{} model, we set the value of $d$ (the size of the learned feature vector) equal to 1024.

The features extracted by each branch of the architecture are combined by concatenation, resulting in a feature vector of 2048 descriptors, i.e., 1024 from the CNN branch and 1024 from the RNN branch. In this way we assure an equal contribution from each branch to the final decision.
Many other combination techniques are possible (e.g., sum, gating, and so on) but we rely on standard concatenation following recent practices introduced by working on multi-source remote sensing analysis~\cite{abs-1803-01945,Gaetano18}. 

\subsection{ Training of \method{} model }
\label{sec:train}

One of the advantages of deep learning approaches, compared to standard machine learning methods, is the ability to link, in a single pipeline, the feature extraction step as well as the associated classifier~\cite{Zhang16}. This ability is particularly important when different flows of information need to be combined together, such as in our scenario where we need to couple different representations of the same data source.
In addition, the different features learned via multiple non-linear combination of the radiometric information are optimized for the specific task at hand, i.e., land cover mapping. 

To further strengthen the complementarity as well as the discriminative power of the learned features for each branch, we adapt the technique proposed in~\cite{HouLW17} to our problem. In~\cite{HouLW17}, the authors propose to learn two complementary representations (using two convolutional networks) from the same image. The discriminative power is enhanced by two auxiliary classifiers, linked to each group of features, in addition to the classifier that uses the merged information. The complementarity is enforced by alternating the optimization of the parameters of the two branches. 
In our case, we still have a unique source of information (an optical Sentinel-2 time series of satellite images) but we manage it via two processing branches that differ from each other regarding the particular deep learning strategy we employ. 

In detail, the classifier that exploits the full set of features is fed by concatenating the output features of both CNN ($cnn_{feat}$) and RNN ($rnn_{feat}$) modules together. Empirically, we have observed that the RNN module overfits the data. To alleviate this problem, we add a Dropout layer~\cite{DahlSH13} on $rnn_{feat}$ with a drop-rate equals to 0.4. 
The learning process involves the optimization of three classifiers at the same time, one specific to $rnn_{feat}$, a second one related to $cnn_{feat}$ and the third one that considers $[rnn_{feat},cnn_{feat}]$. 

The cost function associated to our model is :
\begin{align}
L_{total} &= \alpha_1 * L_1(rnn_{feat}) + \nonumber \\
		  &= \alpha_2 * L_2(cnn_{feat}) + \nonumber \\
          &= L_{fus}([cnn_{feat}, rnn_{feat}]) \label{eqn:cost}
\end{align}

where 

$L_i(feat)$ is the loss function associated to the classifier inputed with the features $feat$. In our case, for all the classifiers (the auxiliary and the main ones) we adopt two fully connected layers of 1024 neurons with ReLU activation function plus a final output layer  with as many neurons as the number of land cover classes to predict. The SoftMax activation function is finally applied~\cite{Zhang16} on the output layer with the aim to produce a kind of probability distribution over the class labels.

Each cost function is modeled through categorical cross entropy, a typical choice for multi-class supervised classification tasks~\cite{IencoGDM17}.

$L_{total}$ is optimized end-to-end. Once the network has been trained, the prediction is carried out only by means of the classifier involving the concatenated features $[cnn_{feat}, rnn_{feat}]$. The cost functions $L_1$ et $L_2$, as highlighted in~\cite{HouLW17}, operate a kind of regularization that forces, within the network, the features extracted to be discriminative independently.

\section{Data}
\label{sec:data}


The analysis was carried out on the \textit{Reunion Island}, a French overseas department located in the Indian Ocean and on part of the  \textit{Gard} department in the South of France. The \textit{Reunion Island} dataset consists of a time series of 34 Sentinel-2 (S2) images acquired between April 2016 and May 2017 while the \textit{Gard} dataset consists of a time series of 37 S2 images acquired between December 2015 and January 2017. All the S2 images used are those provided at level 2A by the THEIA pole~\footnote{Data are available via \url{http://theia.cnes.fr}, preprocessed in surface reflectance via the \textit{MACCS-ATCOR Joint Algorithm}~\cite{Hagolle2015} developed by the National Centre for Space Studies (CNES).}. We only use bands at 10m Blue, Green, Red and Near Infrared (resp. B2, B3, B4 and B8). A preprocessing was performed to fill cloudy observations through a linear multi-temporal interpolation over each band (cf.~\textit{Temporal Gapfilling} \cite{IngladaVATMR17}), and the NDVI radiometric indice was calculated for each date~\cite{IngladaVATMR17,LebourgeoisDVAB17}. A total of 5 variables (4 surface reflectances plus 1 indice) is considered for each pixel of each image in the time series. 
The spatial extent of the \textit{Reunion Island} site is 6\,656 $\times$ 5\,913 pixels while the extent for the  \textit{Gard} site is 4\,822 $\times$ 6\,748 pixels. 
Considering the former benchmark, the field database was built from various sources: (i) the \textit{Registre parcellaire graphique} (RPG)\footnote{RPG is part of the European Land Parcel Identification System (LPIS), provided by the French Agency for services and payment} reference data of 2014, (ii) GPS records from June 2017 and (iii) photo interpretation of the VHSR image conducted by an expert, with knowledge of the territory, for distinguishing between natural and urban spaces. 
Also for the latter benchmark, the field database was built from various sources: (i) the \textit{Registre parcellaire graphique} (RPG)\footnotemark[\value{footnote}] reference data of 2016 and (ii) photo interpretation of the VHSR image conducted by an expert, with knowledge of the territory, for distinguishing between natural and urban areas.

\begin{table}[!ht]
\centering
\begin{tabular}{|l||c|c|c|}
	\hline
\textbf{Class} & Label & \# \textbf{Objects} & \# \textbf{Pixels} \\ 
\hline \hline
1 & {\em Crop Cultivations} & 380 & 12090 \\ \hline
2 & {\em Sugar cane} & 496 & 84136 \\ \hline
3 & {\em Orchards} & 299 & 15477 \\ \hline
4 & {\em Forest plantations} & 67 & 9783 \\ \hline
5 & {\em Meadow} & 257 & 50596 \\ \hline
6 & {\em Forest} & 292 & 55108 \\ \hline
7 & {\em Shrubby savannah} & 371 & 20287 \\ \hline
8 & {\em Herbaceous savannah} & 78 & 5978 \\ \hline
9 & {\em Bare rocks} & 107 & 18659 \\ \hline
10 & {\em Urbanized areas} & 125 & 36178 \\ \hline
11 & {\em Greenhouse crops}& 50 & 1877 \\ \hline
12 & {\em Water Surfaces} & 96 & 7349 \\ \hline
13 & {\em Shadows} & 38 & 5230 \\ \hline
\end{tabular}
\caption{Characteristics of the \textit{Reunion Island} Dataset\label{tab:data_reu}}
\end{table}

\begin{table}[!ht]
\centering
\begin{tabular}{|l||c|c|c|}
	\hline
\textbf{Class} & Label & \# \textbf{Objects} & \# \textbf{Pixels} \\ 
\hline \hline
1 & {\em Corn Crop} & 260 & 115264 \\ \hline
2 & {\em Barley Crop} & 76 & 12279 \\ \hline
3 & {\em Other Crops} & 430 & 127236 \\ \hline
4 & {\em Rice} & 208 & 124943 \\ \hline
5 & {\em Orchards} & 242 & 26142 \\ \hline
6 & {\em Olive Trees} & 203 & 9367 \\ \hline
7 & {\em Meadow} & 230 & 59018 \\ \hline
8 & {\em Vineyard} & 259 & 76815 \\ \hline
9 & {\em Forest} & 171 & 73915 \\ \hline
10 & {\em Urbanized areas} & 266 & 10584 \\ \hline
11 & {\em Water Surfaces} & 193 & 521873 \\ \hline
\end{tabular}
\caption{Characteristics of the Gard Dataset\label{tab:data_gard}}
\end{table}

\subsection{Ground Truth Statistics}
Considering both datasets, ground truth comes in GIS vector file format containing a collection of polygons each attributed with a unique land cover class label. To ensure a precise spatial matching with image data, all geometries have been suitably corrected by hand using the corresponding Sentinel-2 images as reference. Successively, the GIS vector file containing the polygon information has been converted in raster format at the Sentinel-2 spatial resolution (10m).

The final ground truths are constituted of 322\,748 pixels (resp. 2\,656 objects) distributed over 13 classes for the \textit{Reunion Island} dataset (Table~\ref{tab:data_reu}) and 1\,157\,260 pixels (resp. 2\,538 objects) distributed over 8 classes for the \textit{Gard} benchmark (Table~\ref{tab:data_gard}). We remind that the ground truth, in both cases, was collected over large areas. 

\section{Experiments}
\label{sec:expe}
In this section, we present and discuss the experimental results obtained on the study sites introduced in Section~\ref{sec:data}.
We carried out several experimental stages, in order to provide a complete analysis of the behavior of \method{}:
\begin{itemize}
    \item we provide an ablation study in which we evaluate the importance of the different components of \method{} (Section~\ref{sec:abla});
    \item we perform an extensive quantitative analysis, comparing the global classification performances and the per-class results obtained by \method{} w.r.t. competing methods and baseline approaches (Section~\ref{sec:comparison});
    \item we provide a qualitative discussion considering the land cover maps produced by our framework compared to those produced by competing methods (Section~\ref{sec:qualitative}).
\end{itemize}




\subsection{Experimental Settings}

For our analysis, we selected as competing methods two state of the art techniques commonly employed for the classification of SITS. As state of the art standard machine learning tool, we selected a Random Forest ($RF$) classifier~\cite{IngladaVATMR17,LebourgeoisDVAB17}, while as 
regards deep learning techniques
we selected the Recurrent Neural Network approach ($LSTM$) recently proposed in~\cite{IencoGDM17}, since it demonstrates interesting performances in the classification of SITS.
Furthermore, similarly to what proposed in~\cite{IencoGDM17}, we also investigate the possibility to use the deep learning architecture 
to obtain a new data representation for the classification task. To this end, we feed a Random Forest classifier with the features extracted by \method{}, naming this approach $RF(\method)$.

All the Deep Learning methods (including \method{}) are implemented using the Python Tensorflow library. 
During the learning phase, considering both \method{} and $LSTM$, we use the Adam method~\cite{KingmaB14} to learn the model parameters with a learning rate equal to $2 \cdot 10^{-4}$. The training process is conducted over 300 epochs with a batch size equals to 128. The number of hidden units for the RNN module is fixed to 1\,024.

As concerns \method{}, we perform a preprocessing phase in order to associate each pixel to its surrounding area (i.e., to force the learning process to take into account the spatial context).
We consider patches with a spatial extent equals to 5 $\times$ 5, where each patch represents the spatial context of the pixel in position (2,2). This means that for each timestamp we have a cube of information of size (5 $\times$ 5 $\times$ 5), since 5 is the number of raw bands and indices involved in the analysis. Finally, all the patches are associated to a land cover label that corresponds to the label of the central pixel. 
The values are normalized, per band (resp. indices) considering the time series, in the interval $[0,1]$. 
Regarding the $LSTM$ and $RF$ approaches, according to standard literature~\cite{IencoGDM17}, the input is represented by the time series information associated to each pixel.

We divide the dataset into three parts: training, validation and  test set. Training data are used to learn the model while validation data are exploited for model selection varying the parameters of each method. Finally, the model that achieves the best accuracy on the validation set is successively employed to perform the classification on the test set. More in detail, we use 30\% of the objects for the training phase, 20\% of the objects for the validation set while the remaining 50\% are employed for the test phase. We impose that all the pixels of the same object belong exclusively to one of the splits (training, validation or test) to avoid spatial bias in the evaluation procedure~\cite{IngladaVATMR17}. 
Considering the $RF$ models, we optimize the model via two parameters: the maximum depth of each tree and the number of trees in the forest. For the former parameter, we vary it in the range \{20,40,60,80,100\} while for the latter one we takes values in the set \{100, 200, 300,400,500\}.

Experiments are carried out on a workstation with an Intel (R) Xeon (R) CPU E5-2667 v4@3.20Ghz with 256 GB of RAM and four TITAN X GPU. The assessment of the classification performances is done considering global precision (\textit{Accuracy}), \textit{F-Measure}~\cite{IencoGDM17} and \textit{Kappa} measures.

It is known that, depending on the split of the data, the performances of the different methods may vary as simpler or more difficult examples are involved in the training or test set. To alleviate this issue, for each dataset and for each evaluation metric, we report results averaged over ten different random splits performed with the previously presented strategy.

\subsection{Ablation Analysis}
\label{sec:abla}

In this set of experiments we investigate the interplay among the different components of \method{}, setting up an ablation analysis in which we disentangle the benefits of the different parts of our framework. To this end, we compare \method{} with three variants of the original model: i) excluding the use of the auxiliary classifiers (\method{}$_{noAux}$), ii) considering only the Convolutional Neural Network branch ($Cbranch$) and iii) considering only the Recurrent Neural Network branch ($Rbranch$). The results are reported in Table~\ref{tab:abla_reunion} (resp. Table~\ref{tab:abla_gard}) for the \textit{Reunion Island} (reps. \textit{Gard}) study site. 
In both cases we can note that \method{} outperforms the other variants. This fact underlines that: all the different components play a role in the classification process and support our intuition that
combining different models would produce a more diverse and complete representation of the information. The second finding we can underline is that $Cbranch$ and $Rbranch$ behave similarly, showing no real difference in terms of obtained \textit{Accuracy}, \textit{F-Measure} and \textit{Kappa} measure. Even tough the two variants transform the original information differently, the extracted multifaceted knowledge lets them achieve similar performances. Finally, the use of auxiliary classifiers to boost the discriminative power of each branch independently is confirmed to be effective. It can be observed how for both datasets \method{} outperforms the variant without auxiliary classifiers (\method{}$_{noAux}$) as well as the variants involving only one branch of the proposed architecture (i.e., $Cbranch$ and $Rbranch$).

\begin{table}[!ht]
\centering
\scriptsize
 \begin{tabular}{| l || c | c | c |} \hline
& \textit{ Accuracy} & \textit{ F-Measure} &\textit{ Kappa} \\ \hline \hline
 \method{} & 83.72\% $\pm$ 1.08\% & 83.73\% $\pm$ 1.03\% & 0.8089 $\pm$	0.0122 \\ \hline 
 \method{}$_{noAux}$	& 80.28\% $\pm$ 0.68\% & 80.25\% $\pm$ 0.68\%	& 0.7685 $\pm$ 0.0075 \\ \hline 
 $Cbranch$ 	&  79.50\%	$\pm$ 1.00\% & 79.48\% $\pm$ 1.11\%	& 0.7594 $\pm$ 0.0118\\ \hline
 $Rbranch$  & 79.11\% $\pm$ 1.66\% & 78.97\% $\pm$ 1.71\%	& 0.7547 $\pm$ 0.0191	\\ \hline
 \end{tabular}
 \caption{Accuracy, F-Measure, Kappa considering different ablation of \method{} on the \textit{Reunion} dataset \label{tab:abla_reunion}}
\end{table}

\begin{table}[!ht]
\centering
\scriptsize
 \begin{tabular}{| l || c | c | c |} \hline
& \textit{ Accuracy} & \textit{ F-Measure} &\textit{ Kappa} \\ \hline \hline
 \method{} & 96.36\% $\pm$ 0.52\% & 96.28\% $\pm$ 0.55\% & 0.9513 $\pm$ 0.0067\\ \hline 
 \method{}$_{noAux}$	& 95.86\% $\pm$ 0.34\% & 95.78\% $\pm$ 0.33\%	& 0.9446
 $\pm$	0.0041 \\ \hline 
 $Cbranch$ 	&  96.01\% $\pm$  0.42\% & 95.91\%	 $\pm$	0.40\% & 0.9466 $\pm$	0.0056 \\ \hline
 $Rbranch$  & 96.02\%	$\pm$ 0.43\% & 95.91\% $\pm$ 0.40\%	& 0.9466 $\pm$ 0.0056	\\ \hline
 \end{tabular}
 \caption{Accuracy, F-Measure, Kappa considering different ablation of \method{} on the \textit{Gard} dataset \label{tab:abla_gard}}
\end{table}

\subsection{Comparative Analysis}
\label{sec:comparison}

Table~\ref{tab:ReunionGen} and Table~\ref{tab:GardGen} report the results obtained by $RF$, $LSTM$ , \method{} and $RF(\method)$
on the \textit{Reunion Island} and \textit{Gard} study sites, respectively.
We can observe how on both benchmarks \method{} outperforms both state of the art competing methods. However, we can also note that using \method{} as feature extractor ($RF(\method)$) provides always the best average performances in terms of \textit{Accuracy}, \textit{F-Measure} and \textit{Kappa} measure. This is in line with recent work in remote sensing~\cite{Dinh18,abs-1806-11452} and it is due to the fact that, once the newly extracted features are optimized for a particular task, classical machine learning techniques are able to efficiently leverage the information richness carried out by such features.

\begin{table}[!ht]
\centering
\scriptsize
 \begin{tabular}{| l || c | c | c |} \hline
& \textit{ Accuracy} & \textit{ F-Measure} &\textit{ Kappa} \\ \hline \hline
 \textit{RF} & 82.99\% $\pm$ 1.04\% & 82.40\%	$\pm$ 1.09\% & 0.7989 $\pm$	0.0119 \\ \hline
 \textit{LSTM} & 76.66\%	$\pm$  1.21\%  & 76.57\% $\pm$ 1.11\% &  0.7260 $\pm$  0.0140 \\ \hline
 \method{} & 83.72\% $\pm$ 1.08\% & 83.73\% $\pm$ 1.03\% & 0.8089 $\pm$	0.0122 \\ \hline 
 \textit{RF(\method{})} & \textbf{86.12}\% $\pm$ 1.21\% & \textbf{86.00}\%	$\pm$ 1.24\% & \textbf{0.8366} $\pm$	0.0143 \\ \hline 
\hline
 \end{tabular}
 \caption{REUNION \label{tab:ReunionGen}}
\end{table}

\begin{table}[!ht]
\centering
\scriptsize
 \begin{tabular}{| l || c | c | c |} \hline
& \textit{ Accuracy} & \textit{ F-Measure} &\textit{ Kappa} \\ \hline \hline
 
 \textit{RF} & 96.04\%	$\pm$ 0.40\% & 95.71\% $\pm$ 0.44\% & 0.9469
 $\pm$  0.0046 \\ \hline 
 \textit{LSTM} & 95.05\% $\pm$ 0.55\% & 94.81\% $\pm$ 0.59\% & 0.9338 $\pm$ 0.0066 \\ \hline
 \method{} & 96.36\% $\pm$ 0.52\% & 96.28\% $\pm$ 0.55\% & 0.9513 $\pm$ 0.0067\\ \hline 
 \textit{RF(\method{})} & \textbf{96.78}\% $\pm$ 0.50\% & \textbf{96.70}\% $\pm$ 0.51\% & \textbf{0.9569} $\pm$ 0.0061\\ \hline 
\hline
 \end{tabular}
 \caption{GARD \label{tab:GardGen}}
\end{table}

\subsubsection{Per-Class Analysis on the \textit{Reunion Island} benchmark}
Table~\ref{tab:PerClass_fm_reunion} and Table~\ref{tab:PerClass_fm_gard} summarize the  per class \textit{F-Measure} performances of the different methods for the \textit{Reunion Island} and \textit{Gard} study site, respectively.

\begin{figure}[!ht]
\centering
\subfloat[\label{fig:cm_rf_reunion}] {\includegraphics[width=.49\linewidth]{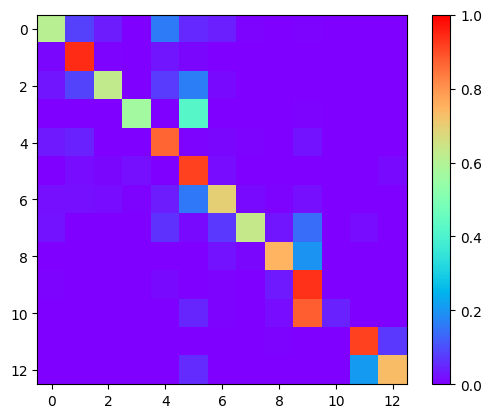} }
\subfloat[\label{fig:cm_lstm_reunion}] {\includegraphics[width=.49\linewidth]{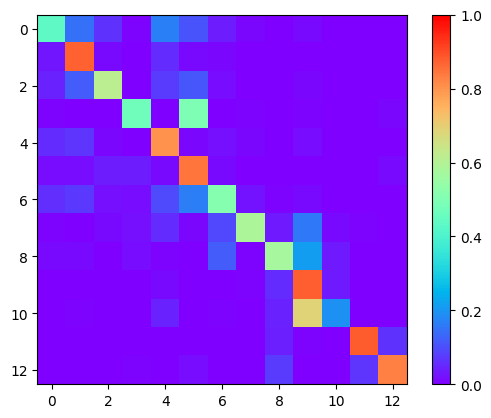} }\\
\subfloat[\label{fig:cm_deep_reunion}] {\includegraphics[width=.49\linewidth]{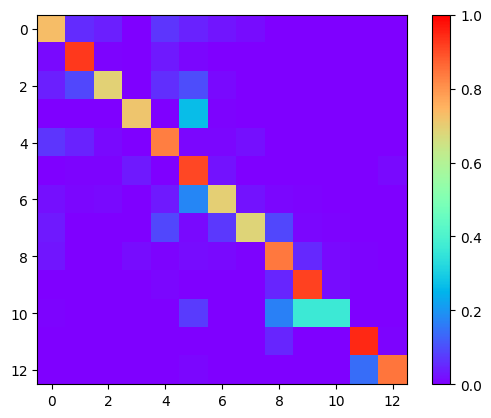}}
\subfloat[\label{fig:cm_rf_deep_reunion}]
{\includegraphics[width=.49\linewidth]{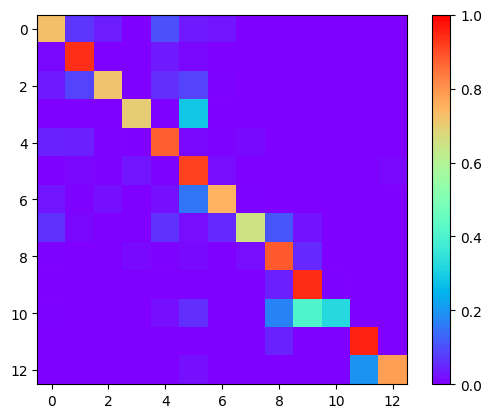}}
\caption{ Heat Maps representing the confusion matrices of: (a) $RF$, (b) $LSTM$, (c) \method{} and (d) $RF(\method)$ on the \textit{Reunion Island} study site. \label{fig:cm_reunion}}
\end{figure}

Considering the \textit{Reunion Island} benchmark (Table~\ref{tab:PerClass_fm_reunion}), we can observe that, considering the main competing approaches ($RF$, $LSTM$ and \method), our framework supplies the best classification results on nine over thirteen land cover classes. These classes are: (0), (1), (2), (3), (8), (9), (10), (11) and (12) (resp. \textit{Crop Cultivations}, \textit{Sugar cane}, \textit{Orchards}, \textit{Forest plantations}, \textit{Herbaceous Savannah}, \textit{Bare rocks}, \textit{Urbanized areas}, \textit{Greenhouse crops}, \textit{Water surfaces} and \textit{Shadows}). The highest gains are obtained in correspondence of \textit{Bare rocks}, \textit{Urbanized areas} and \textit{Greenhouse crops} classes with an improvement of 8, 9 and 20 point of \textit{F-Measure}, respectively. Considering the characteristics of such classes, the significant gains obtained by \method{} are the results of the effectiveness of our approach to exploit the temporal behavior supplied by the time series as well as the fact that our approach integrates a small amount of spatial context via the patch (a 5 $\times$ 5 image grid) centered around the analyzed pixel. Regarding all the other land cover classes (\textit{Forest Plantations}, \textit{Meadow}, \textit{Forest} and \textit{Shrubby savannah}) the performances of \method{} are comparable with the results obtained by the other approaches. Looking at the $RF(\method)$ method, we can observe that this solution provides the best performances over all the land cover categories and, in particular, we can note that the $RF$ algorithm largely benefits from the feature extracted by the deep learning architecture increasing its average results of more than 3 points.
We also investigate the confusion between each pair of classes and we report, for each competing method, the confusion matrix in Figure~\ref{fig:cm_reunion}. The visual results support the previous discussion, since a closer look at the heat maps representing the confusion matrix points out that \method{} and $RF(\method)$ are more precise than the competitors. This consideration emerges from the fact that the corresponding heat maps (Figure~\ref{fig:cm_deep_reunion} and Figure~\ref{fig:cm_rf_deep_reunion}) have a more visible diagonal structure (the dark red blocks concentrated on the diagonal). This is not the case for Random Forest (Figure~\ref{fig:cm_rf_reunion}) and $LSTM$ (Figure~\ref{fig:cm_lstm_reunion}) where the distinction between different classes is less sharp.

\begin{table}[!ht]
\large
\centering
\begin{adjustbox}{width=1.\textwidth}
\begin{tabular}{|l||c|c|c|c|c|c|c|c|c|c|c|c|c|}
	\hline
\textbf{Method} & 0 & 1 & 2 & 3 & 4 & 5 & 6 & 7 & 8 & 9 & 10 & 11 & 12\\  \hline \hline

\textit{RF} & 61.67\%	& 91.94\% & 70.12\%	& 65.63\% & 83.10\%	& 85.91\%	& 73.23\%	& 67.47\%	& 73.96\%	& 82.98\%	& 10.87\%	& 92.53\%	& 88.40\% \\ \hline
\textit{LSTM} & 42.68\%	& 88.20\%	& 64.20\%	& 53.56\%	& 76.51\%	& 79.51\%	& 59.01\%	& 60.53\%	& 70.86\%	& 81.61\%	& 18.23\%	& 92.16\%	& 86.55\% \\ \hline
\method{} & 62.36\%	& 92.09\% & 73.24\%	& 70.40\% & 82.88\%	& 84.59\% & 70.29\% & 63.40\%	& 82.02\% & 90.47\%	& 40.31\% & 93.26\%	& \textbf{90.76}\% \\ \hline

\textit{RF(\method{})} & \textbf{65.72}\%	& \textbf{92.98}\%	& \textbf{75.39}\%	& \textbf{73.22}\%	& \textbf{85.40}\%	& \textbf{87.30}\%	& \textbf{75.76}\% &	\textbf{67.97}\%	& \textbf{86.32}\%	& \textbf{92.05}\%	& \textbf{43.88}\%	& \textbf{93.87}\%	& 90.29\% \\ \hline
\hline
\end{tabular}
\end{adjustbox}
\caption{Average \textit{ F-Measure} per class for the \textit{Reunion Island} Dataset\label{tab:PerClass_fm_reunion}}
\end{table}

\subsubsection{Per-Class Analysis on the \textit{Gard} benchmark}
Considering the \textit{Gard} benchmark (Table~\ref{tab:PerClass_fm_gard}), we can observe that all the main competing methods achieve similar performances on all the land cover classes with the exception of the \textit{Barley Crop} land cover category. While the results on all other land cover classes are satisfactory regarding all the approaches, conversely, on this class we can observe a clear different behavior between \method{} and the other algorithms. $RF$ and $LSTM$ have a poor behavior on \textit{Barley Crop} with a maximum (average) \textit{F-Measure} of 31.55\%. On the other hand, we can underline that our strategy is capable to reach an \textit{F-Measure} of 59.30\% with an increase of more than 27 percentage points compared to the best state of the art method. Inspecting deeply the results, we have seen that confusion arises between  the \textit{Corn Crop} and \textit{Barley Crop} classes. This is due to a very similar behavior between such classes, which makes it difficult to discriminate between them. The ability of \method{} to disentangle and characterize the temporal behavior of each class efficiently supports the classification of such classes. 
We can also observe that the $RF(\method)$ strategy provides the best performances over all the land cover categories with the exception of the \textit{Forest} class, where its performance is similar to the one of the best approach, namely \method{} ($97.60\%$ vs $97.71\%$).

\begin{figure}[!ht]
\centering
\subfloat[\label{fig:cm_rf_gard}] {\includegraphics[width=.49\linewidth]{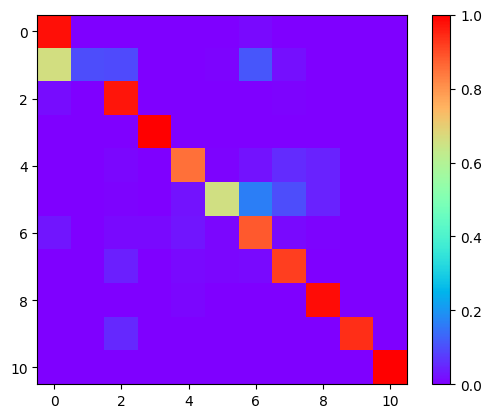} }
\subfloat[\label{fig:cm_lstm_gard}] {\includegraphics[width=.49\linewidth]{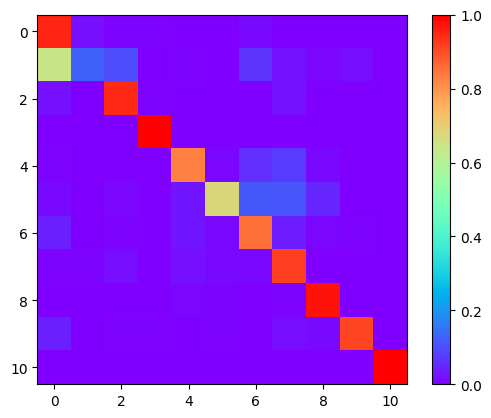} } \\
\subfloat[\label{fig:cm_deep_gard}] {\includegraphics[width=.49\linewidth]{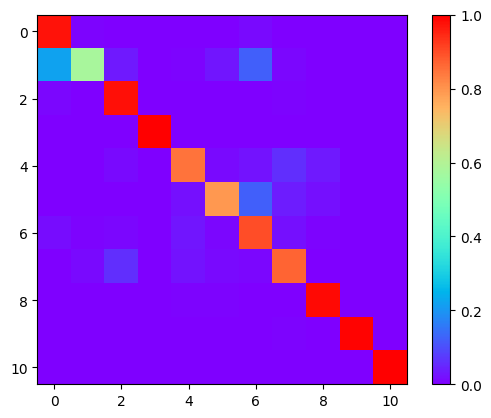}}
\subfloat[\label{fig:cm_rf_deep_gard}]
{\includegraphics[width=.49\linewidth]{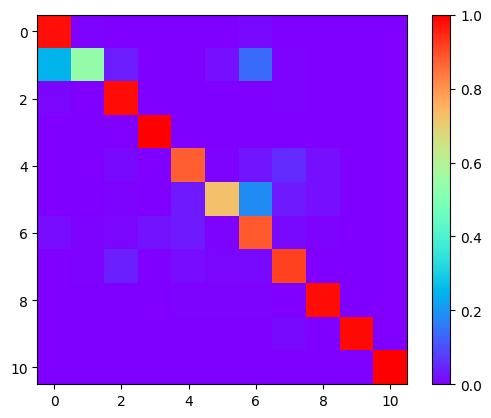}}

\caption{Heat Maps representing the confusion matrices of: (a) $RF$, (b) $LSTM$, (c) \method{} and (d) $RF(\method)$ on the \textit{Gard} study site.\label{fig:cm_gard}}
\end{figure}

Figure~\ref{fig:cm_gard} shows the heat maps representing the confusion matrices on the \textit{Gard} study site. Also in this case, the visual results support the previous discussion. The heat maps representing the confusion matrices pinpoint that \method{} and $RF(\method)$ avoid confusion on some particular classes on which the competitors fail. More specifically, we can observe this phenomenon in two different scenarios: the first one is related to the confusion between the \textit{Corn Crop} and the \textit{Barley Crop} land cover classes while the second case involves the \textit{Olive Trees} class. In the former case, \method{} and $RF(\method)$ are more precise on the \textit{Barley Crop}  while in the latter case, our approaches make some confusion between the \textit{Olive Trees} and the \textit{Meadow}, while the other approaches confuse the \textit{Olive Trees} class not only with \textit{Meadow}, but also with \textit{Vineyard} and \textit{Forest}.

\begin{table}[!ht]
\large
\centering
\begin{adjustbox}{width=1.\textwidth}
\begin{tabular}{|l||c|c|c|c|c|c|c|c|c|c|c|}
	\hline
\textbf{Method} & 0 & 1 & 2 & 3 & 4 & 5 & 6 & 7 & 8 & 9 & 10 \\  \hline \hline

\textit{RF} & 93.40\% & 31.55\%	& 96.33\% & 98.69\%	& 81.19\% & 72.70\%	& 82.63\%	& 92.53\% & 96.74\% & 97.15\%	& 99.80\% \\ \hline
\textit{LSTM} & 91.30\% & 27.58\% & 95.81\% & 98.48\% & 75.37\% & 69.93\% & 78.95\% & 89.48\% & 96.72\% & 93.60\% & 99.82\% \\ \hline
\method{} & 94.92\% &	59.30\% & 96.90\% &	98.99\%	& 80.47\%&  70.93\% & 83.28\% &	91.69\% &	\textbf{97.71}\% &	96.77\% & 99.81\% \\ \hline
\textit{RF(\method{})} & \textbf{95.26}\%	& \textbf{62.65}\%	& \textbf{97.43}\%	& \textbf{99.10}\% &	\textbf{82.18}\%	 & \textbf{74.64}\% & \textbf{84.66}\% & \textbf{93.39}\%	& 97.60\%	& \textbf{98.63}\%	& \textbf{99.86}\% \\ \hline
\hline
\hline
\end{tabular}
\end{adjustbox}
\caption{Average \textit{ F-Measure} per class for the Gard Dataset\label{tab:PerClass_fm_gard}}
\end{table}

\subsection{Qualitative Inspection of Land Cover Maps}
\label{sec:qualitative}

\begin{figure*}[!ht]
\centering
\begin{tabular}{cccc}
\textit{VHSR Image} & \textbf{$RF$} & \textbf{$LSTM$} & \textbf{ \method } \\ 
\subfloat[\label{fig:gard_ex1_b}] {\includegraphics[width=.22\linewidth]{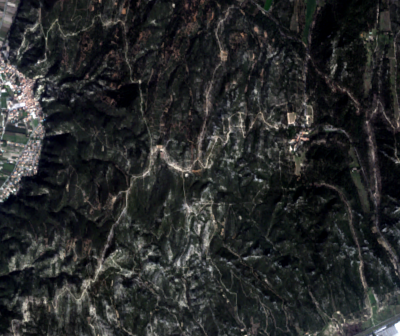} }
&
\subfloat[\label{fig:gard_ex1_rf}] {\includegraphics[width=.22\linewidth]{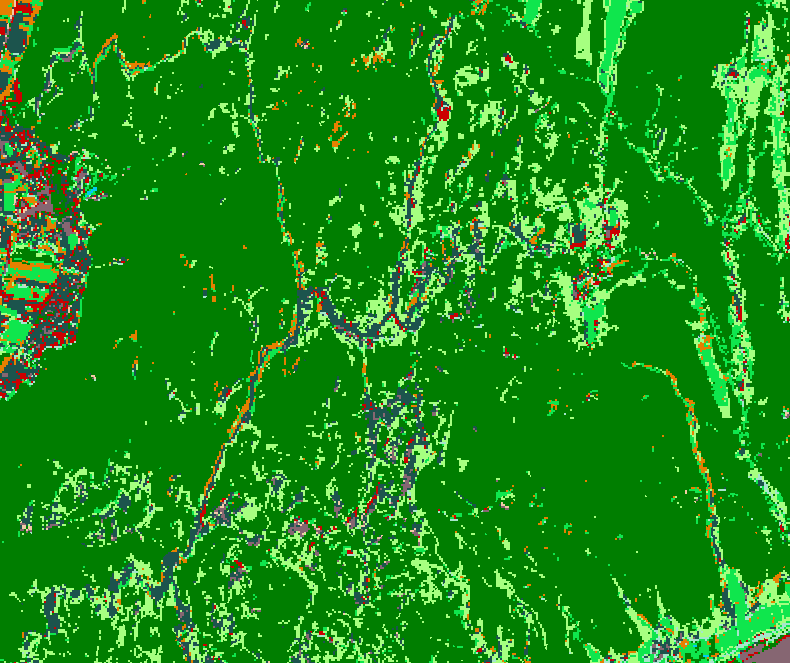} }
&
\subfloat[\label{fig:gard_ex1_lstm}] {\includegraphics[width=.22\linewidth]{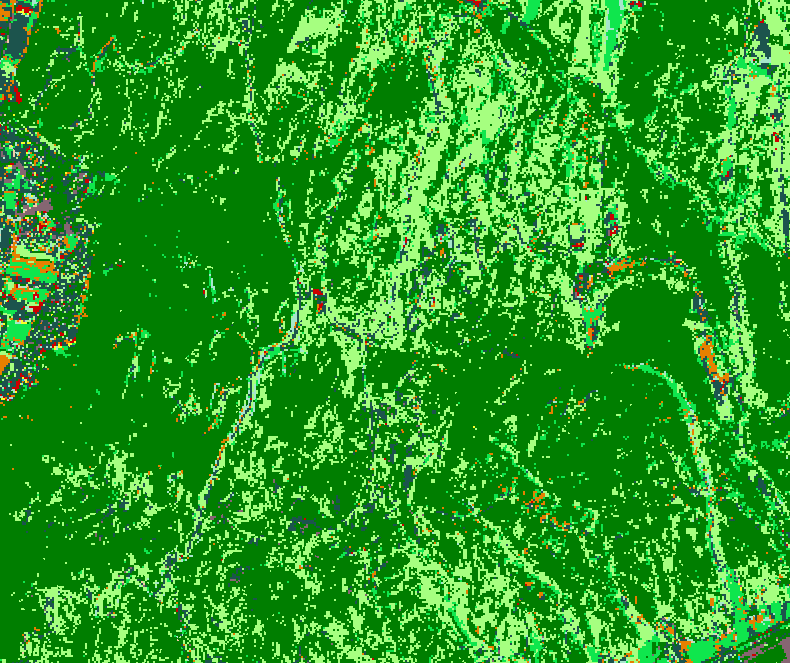} }
&
\subfloat[\label{fig:gard_ex1_duplo}] {\includegraphics[width=.22\linewidth]{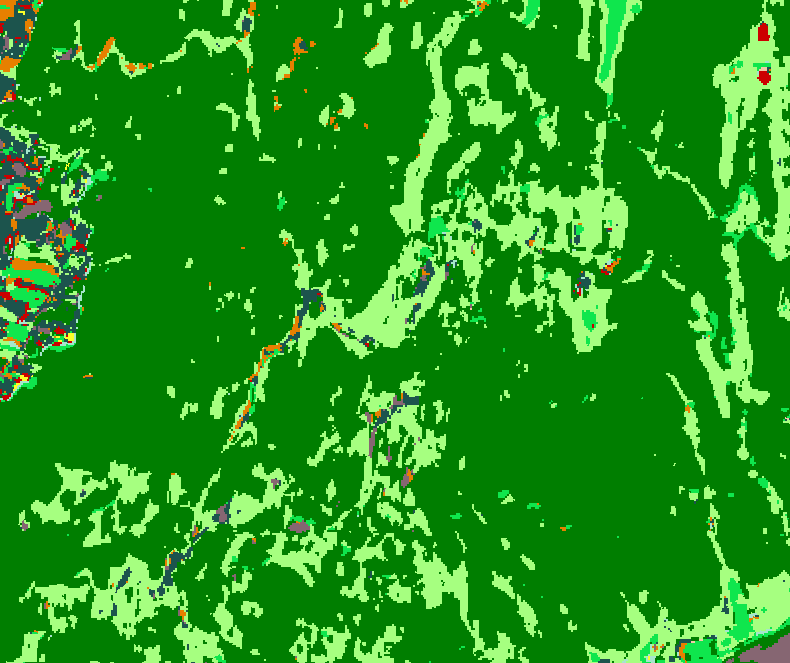} }
\\  
\subfloat[\label{fig:gard_ex2_b}] {\includegraphics[width=.22\linewidth]{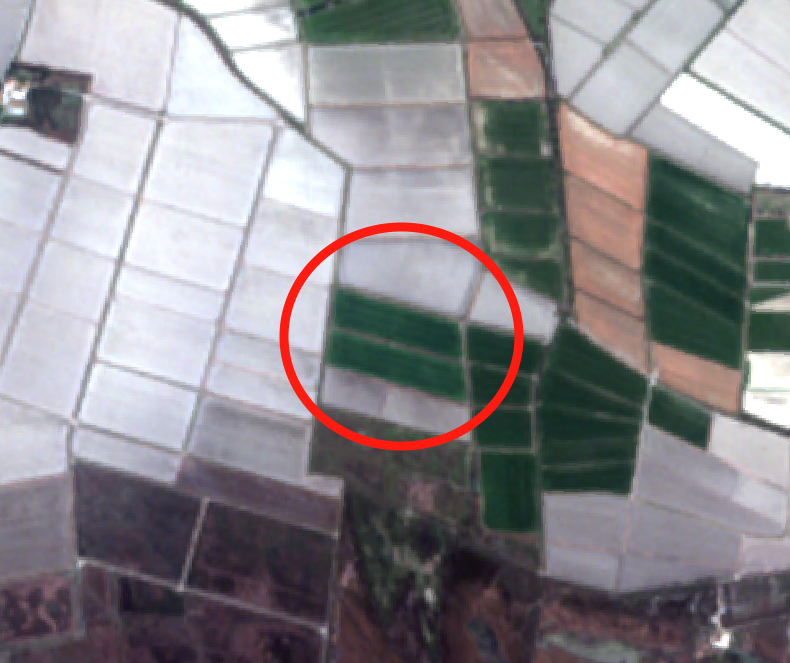} }
&
\subfloat[\label{fig:gard_ex2_rf}] {\includegraphics[width=.22\linewidth]{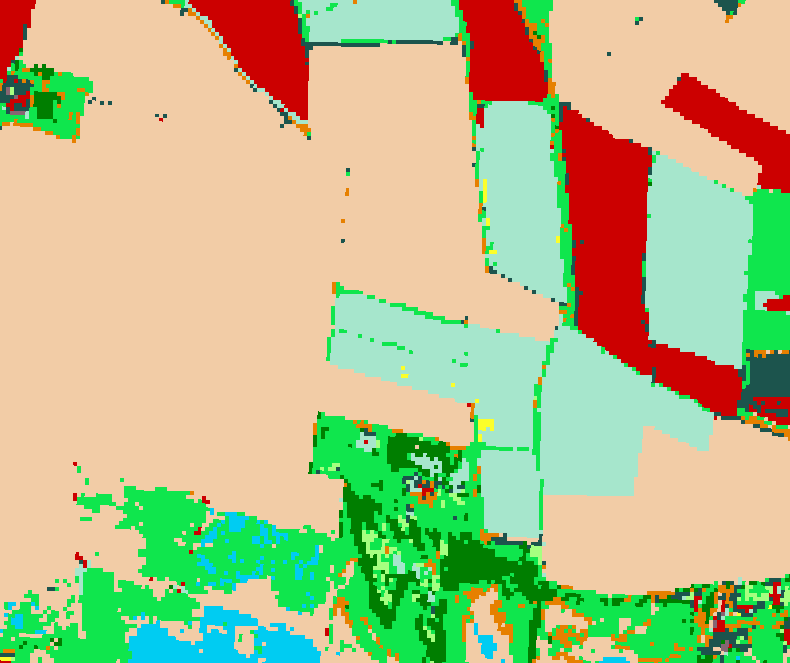} }
&
\subfloat[\label{fig:gard_ex2_lstm}] {\includegraphics[width=.22\linewidth]{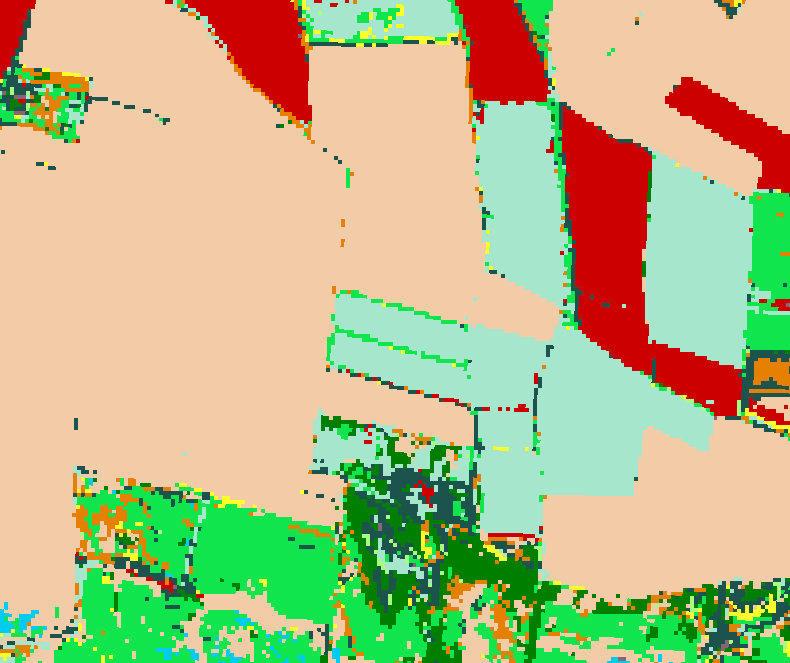} }
&
\subfloat[\label{fig:gard_ex2_duplo}] {\includegraphics[width=.22\linewidth]{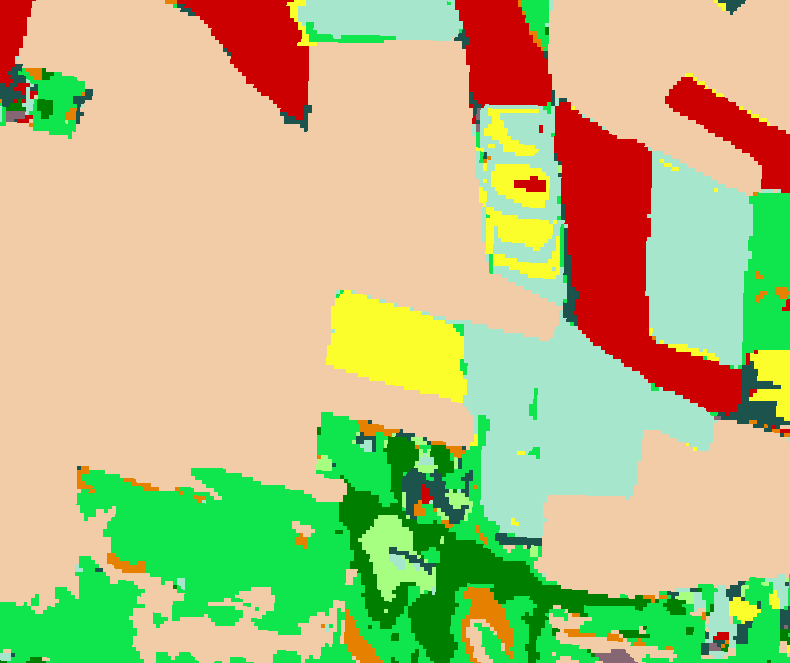} }

\\
\subfloat[\label{fig:gard_ex3_b}] {\includegraphics[width=.22\linewidth]{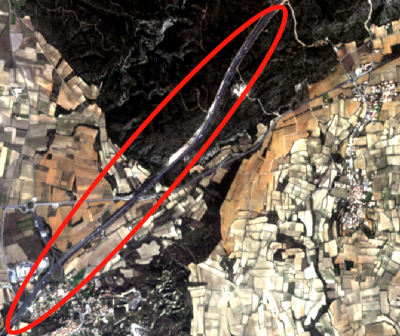} }
&
\subfloat[\label{fig:gard_ex3_rf}] {\includegraphics[width=.22\linewidth]{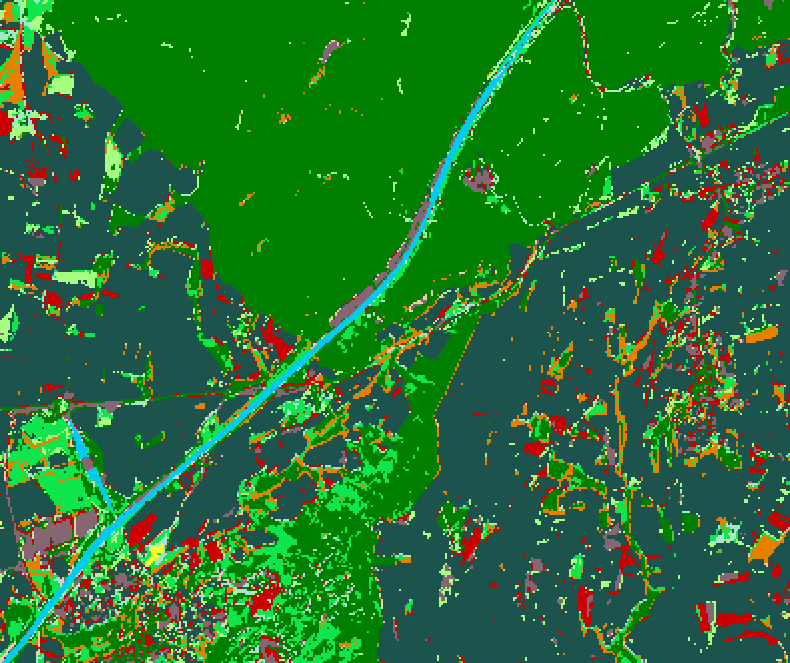} }
&
\subfloat[\label{fig:gard_ex3_lstm}] {\includegraphics[width=.22\linewidth]{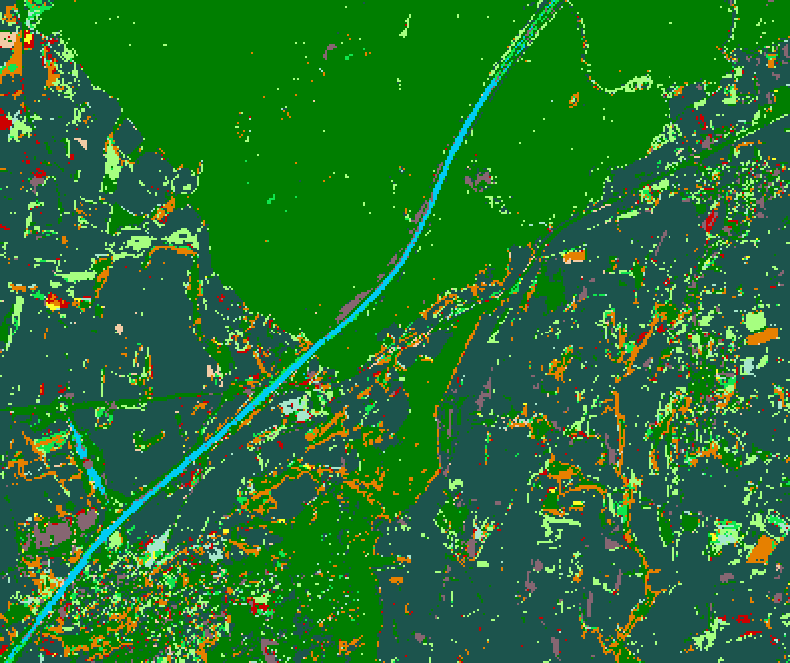} }
&
\subfloat[\label{fig:gard_ex3_duplo}] {\includegraphics[width=.22\linewidth]{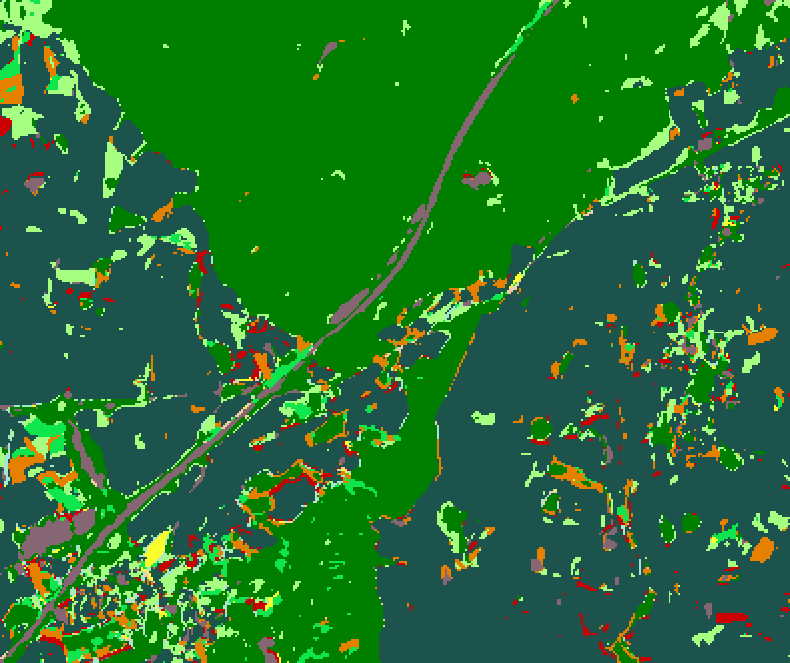} }
\\ 
\multicolumn{4}{c}{ \includegraphics[width=1.\textwidth]{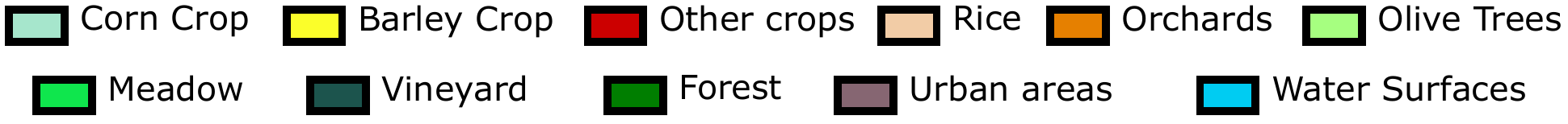}  }

\end{tabular}
\caption{Qualitative investigation of Land Cover Map details produced on the \textit{Gard} study site by\textit{RF}, \textit{LSTM} and \method{}  on three different zones (from the top to the bottom): i) mixed area (forest, meadows and olive trees); ii) rural area and iii) mixed area (urban, rural area and forest).~\label{tab:gard_examples}}
\end{figure*}

\begin{figure*}[!ht]
\centering
\begin{tabular}{cccc}
\textit{VHSR Image} & \textbf{$RF$} & \textbf{$LSTM$} & \textbf{ \method } \\ 
\subfloat[\label{fig:reunion_ex1_b}] {\includegraphics[width=.22\linewidth]{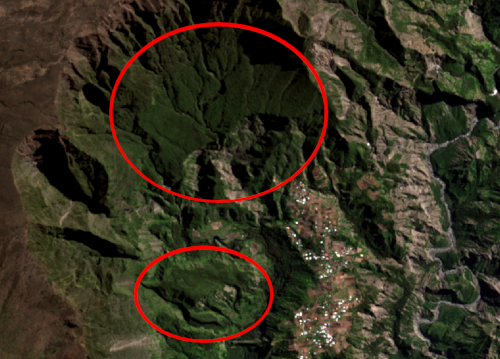} }
&
\subfloat[\label{fig:reunion_ex1_rf}] {\includegraphics[width=.22\linewidth]{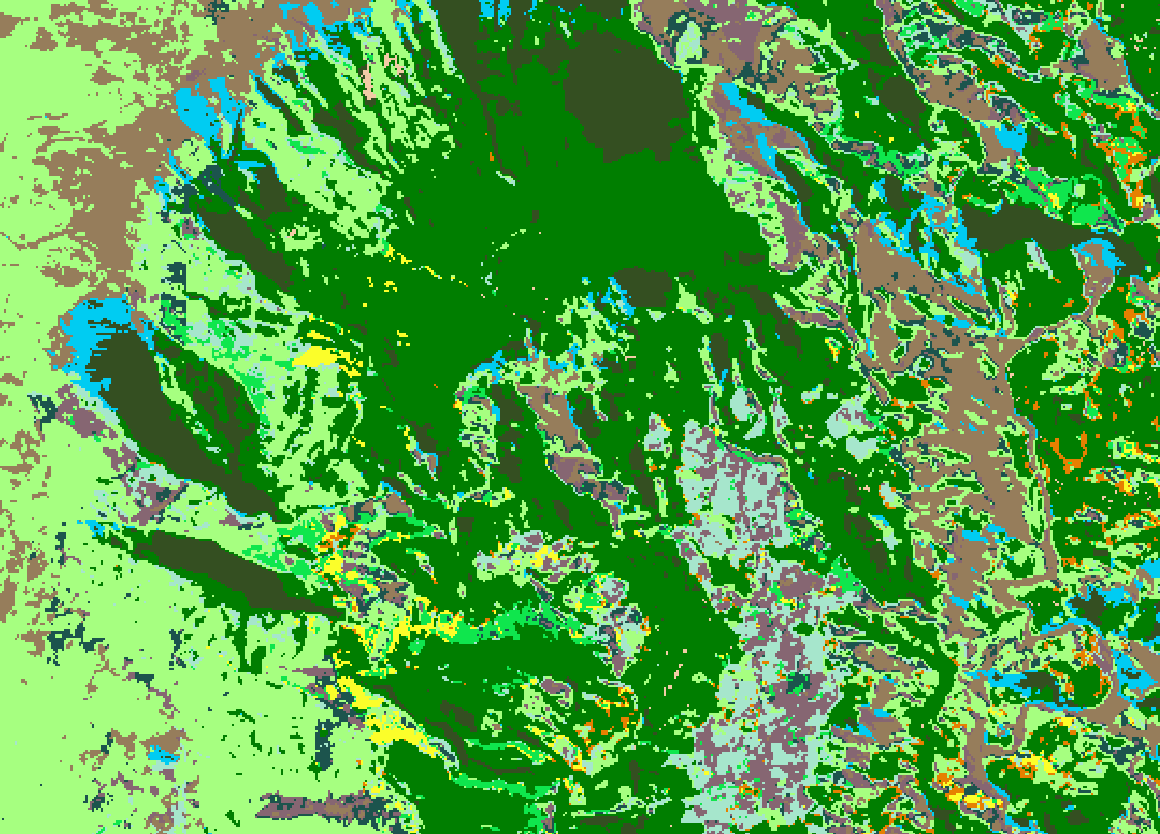} }
&
\subfloat[\label{fig:reunion_ex1_lstm}] {\includegraphics[width=.22\linewidth]{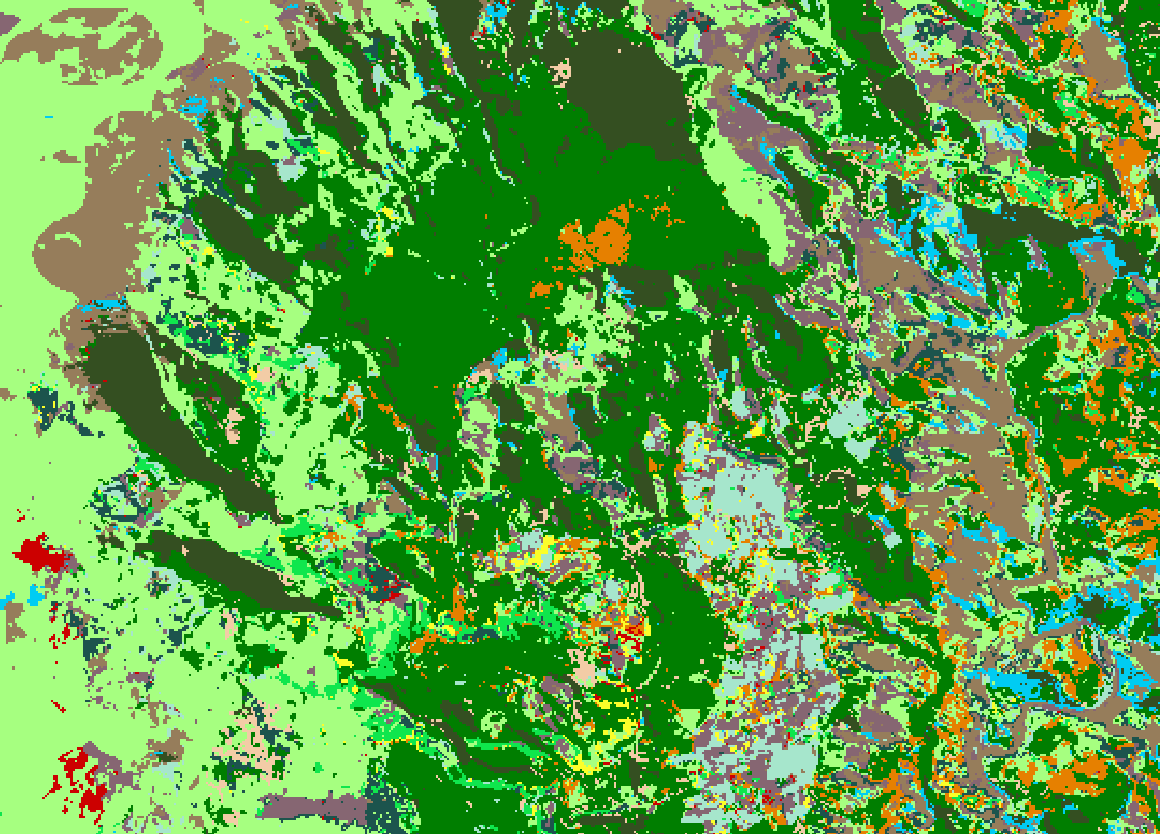} }
&
\subfloat[\label{fig:reunion_ex1_duplo}] {\includegraphics[width=.22\linewidth]{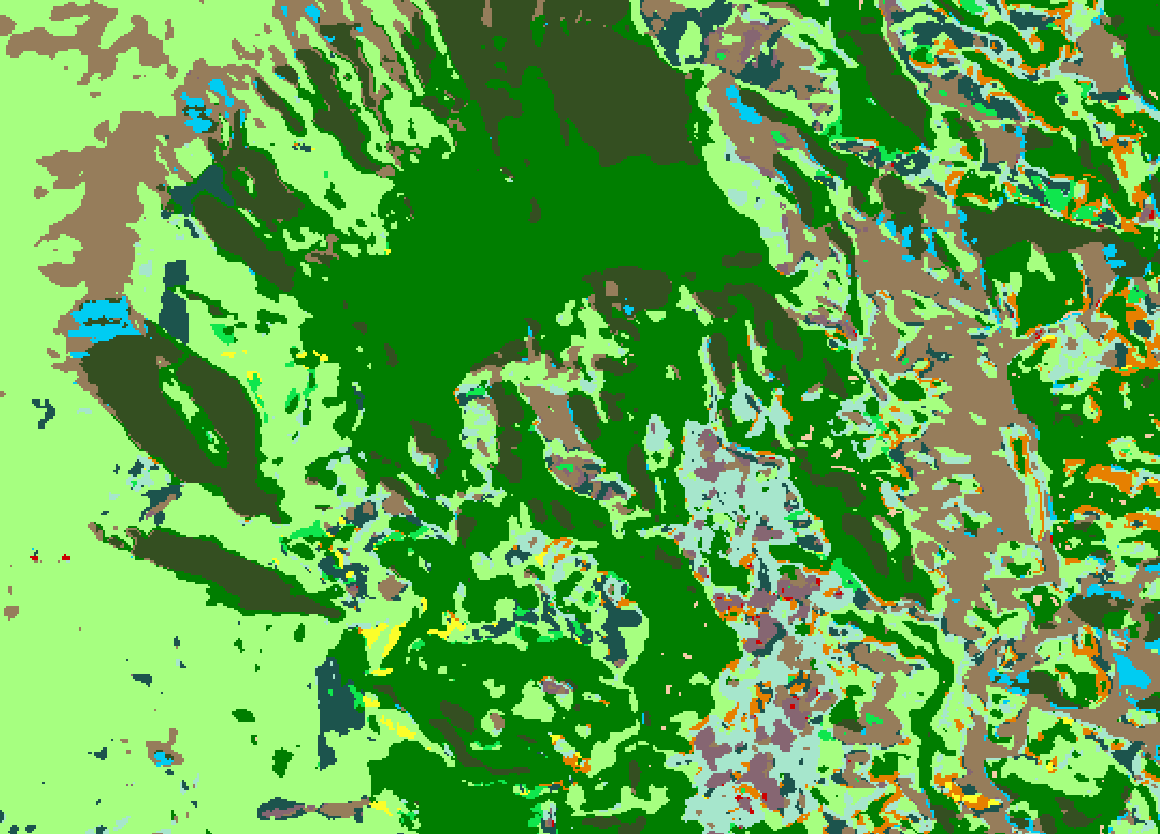} }
\\  
\subfloat[\label{fig:reunion_ex2_b}] {\includegraphics[width=.22\linewidth]{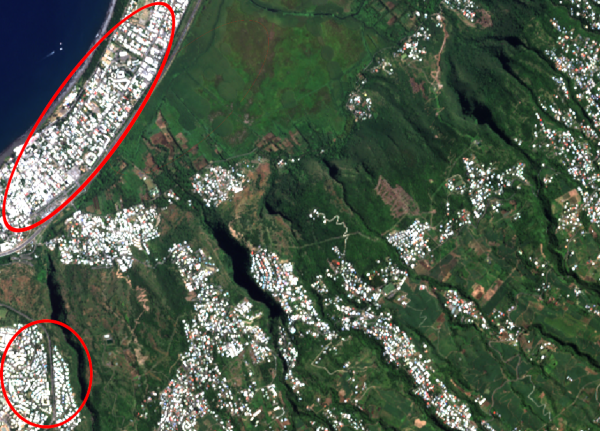} }
&
\subfloat[\label{fig:reunion_ex2_rf}] {\includegraphics[width=.22\linewidth]{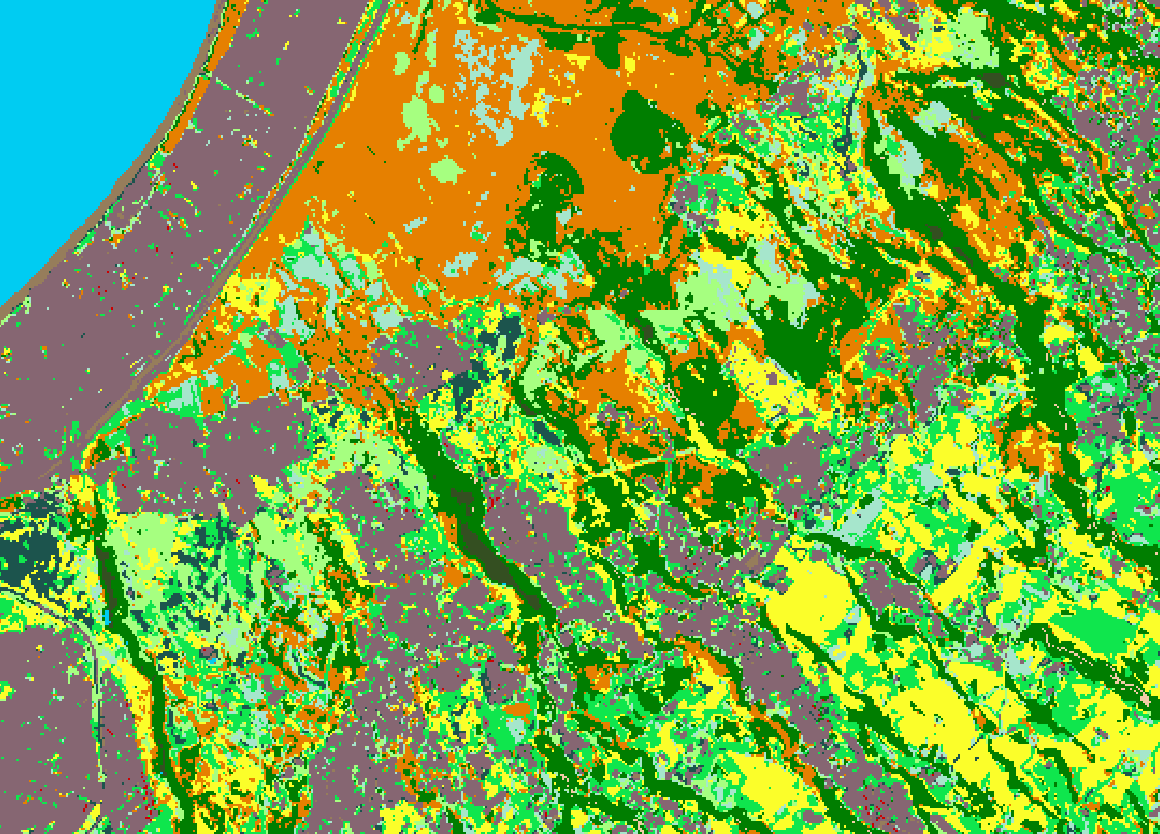} }
&
\subfloat[\label{fig:reunion_ex2_lstm}] {\includegraphics[width=.22\linewidth]{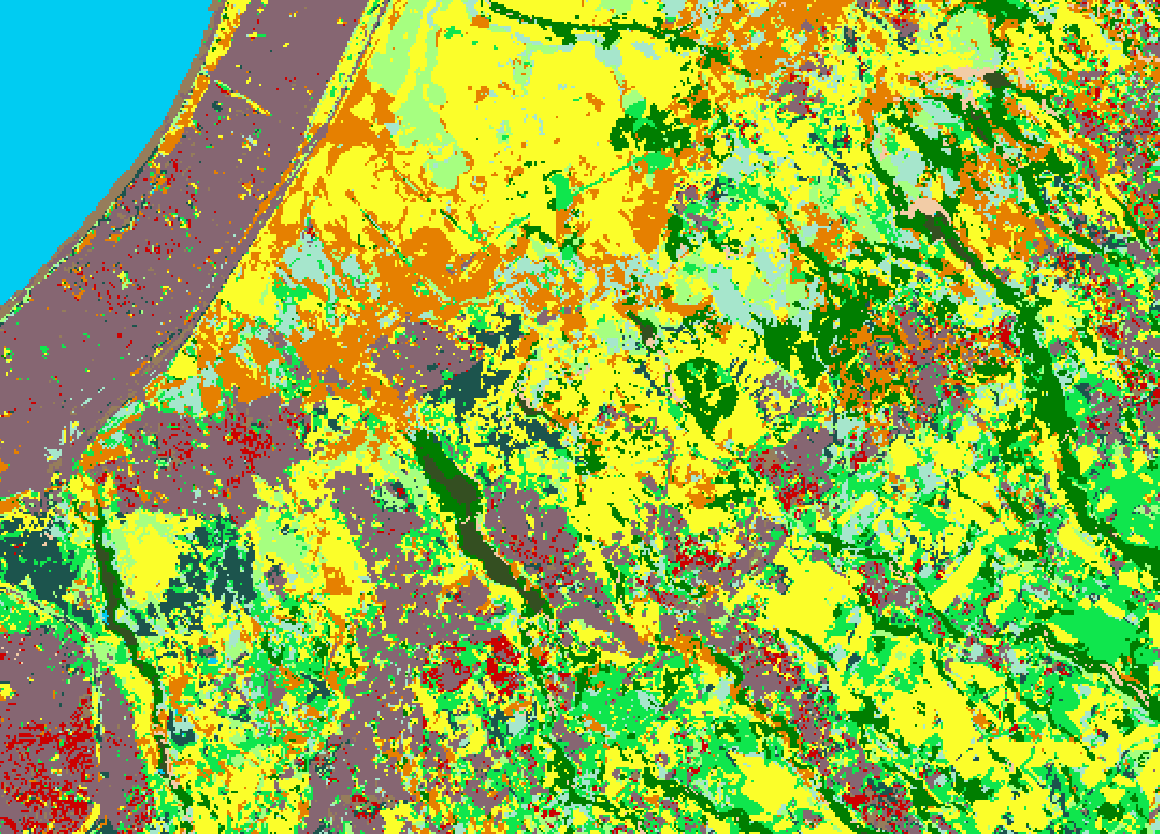} }
&
\subfloat[\label{fig:reunion_ex2_duplo}] {\includegraphics[width=.22\linewidth]{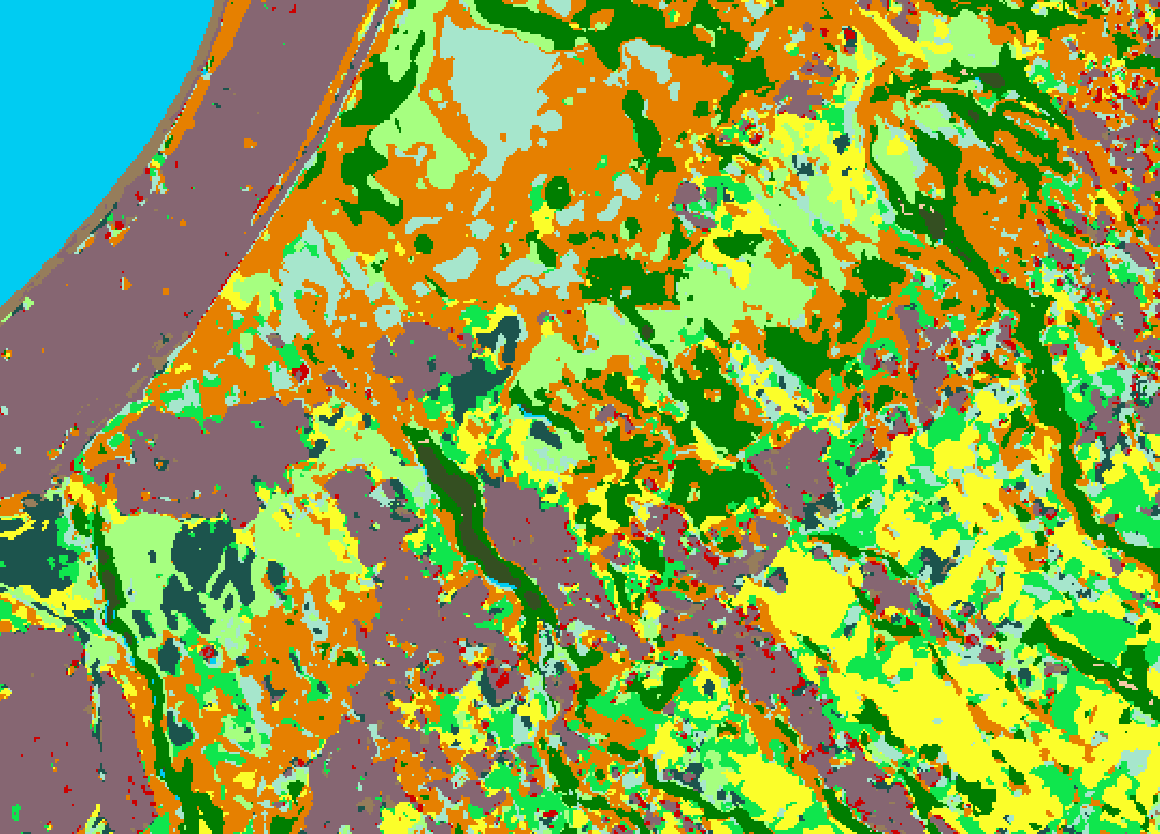} }

\\
\subfloat[\label{fig:reunion_ex3_b}] {\includegraphics[width=.22\linewidth]{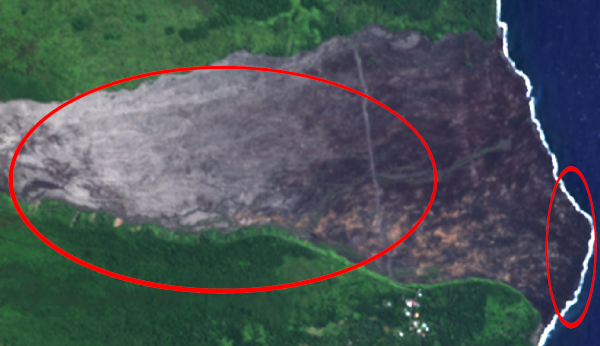} }
&
\subfloat[\label{fig:reunion_ex3_rf}] {\includegraphics[width=.22\linewidth]{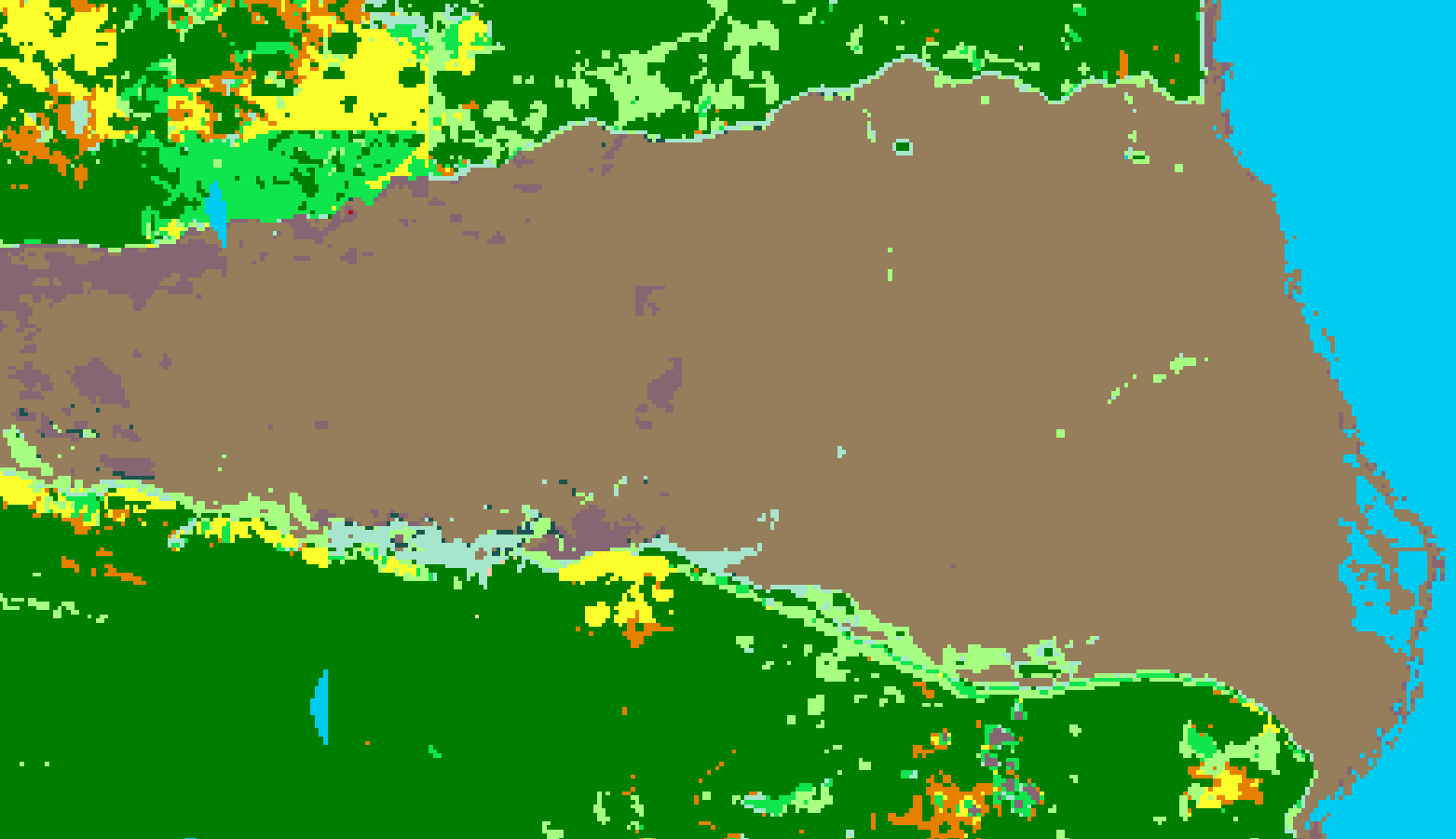} }
&
\subfloat[\label{fig:reunion_ex3_lstm}] {\includegraphics[width=.22\linewidth]{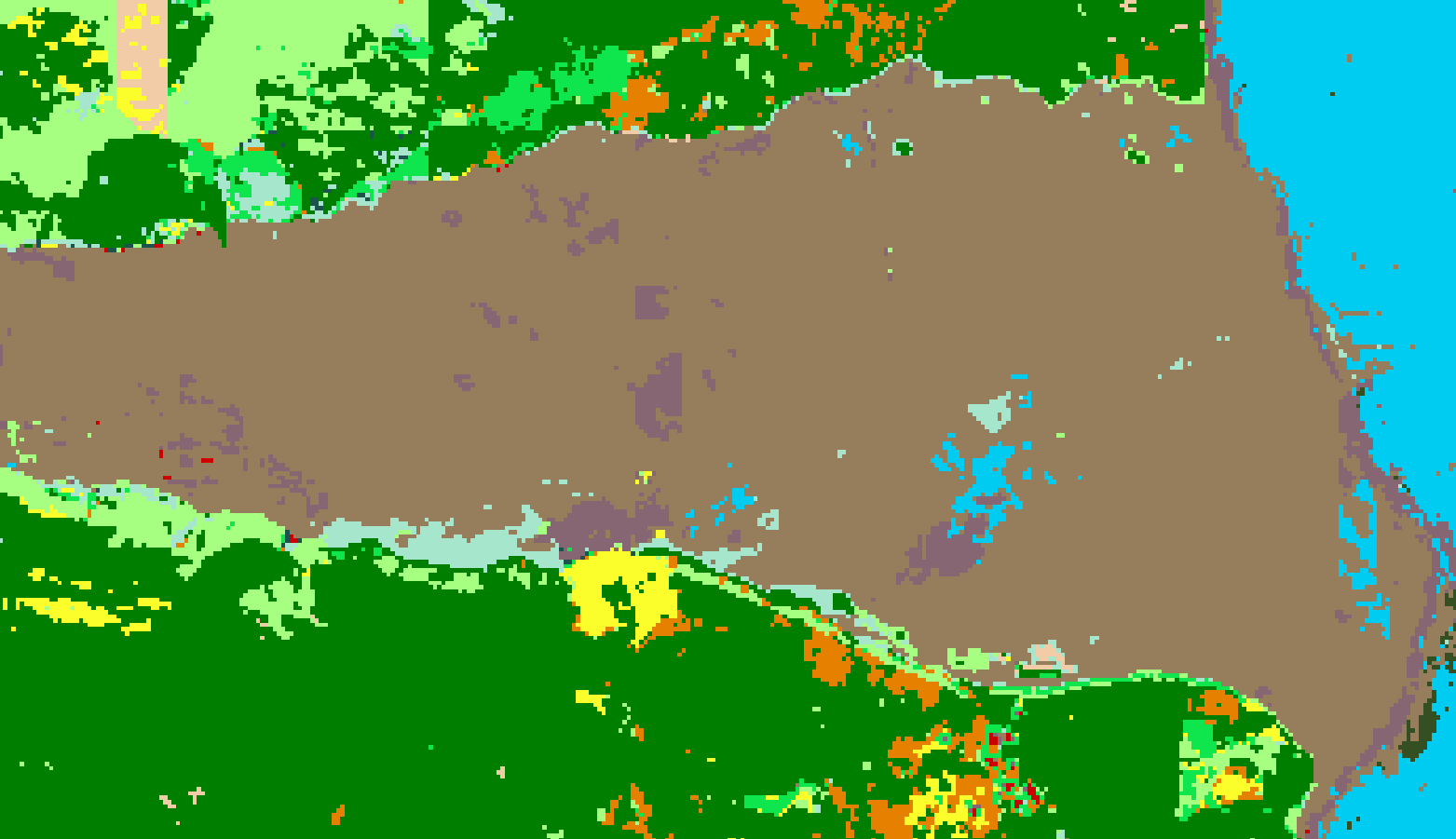} }
&
\subfloat[\label{fig:reunion_ex3_duplo}] {\includegraphics[width=.22\linewidth]{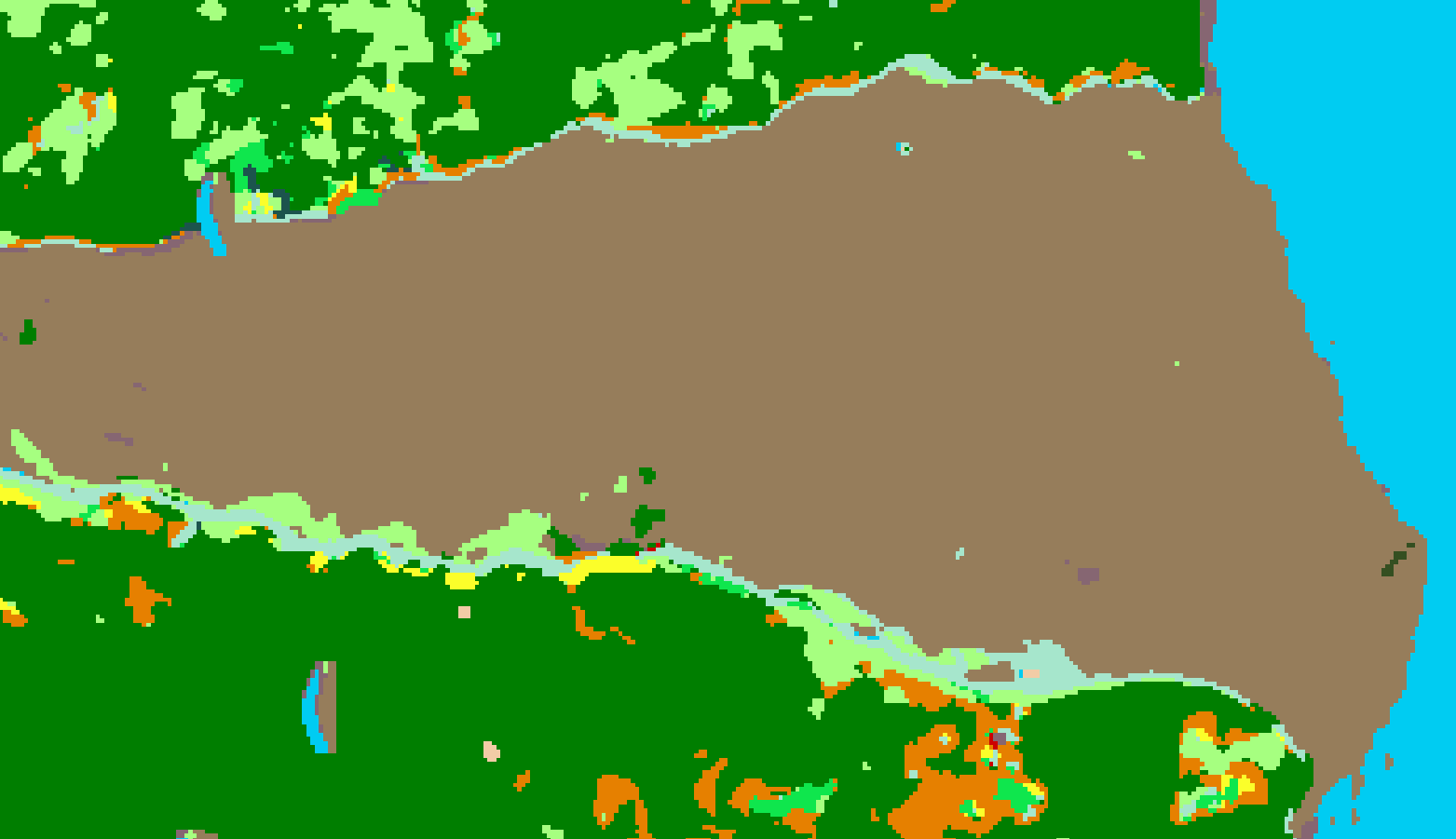} }
\\ 
\multicolumn{4}{c}{ \includegraphics[width=1.\textwidth]{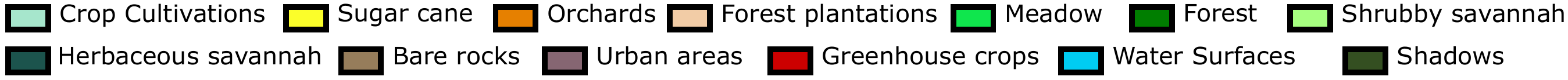}  }

\end{tabular}
\caption{Qualitative investigation of Land Cover Map details produced on the \textit{Reunion Island} study site by\textit{RF}, \textit{LSTM} and \method{}  on three different zones (from the top to the bottom): i) forest area; ii) urban area and iii) bare rocks.~\label{tab:reunion_examples}}
\end{figure*}

In Figure~\ref{tab:gard_examples} and~\ref{tab:reunion_examples} we report some representative map classification details on the \textit{Gard} and \textit{Reunion Island} datasets considering the \textit{RF}, \textit{LSTM} and \method, respectively. With the purpose to supply a reference image with natural colors for the map classification details, we have used the multispectral information obtained from a Very High Spatial Resolution image acquired on the same area in the interval spanned by the time series. More in detail, for each study site, we used the multispectral bands of a SPOT7 image at a spatial resolution of 6m.

Regarding the~\textit{Gard} study site, the first example (Figures~\ref{fig:gard_ex1_b}, \ref{fig:gard_ex1_rf}, \ref{fig:gard_ex1_lstm} and~\ref{fig:gard_ex1_duplo}) depicts an area mainly characterized by forest, meadows and olive tress. On this area, we can observe that both $RF$ and $LSTM$ present confusion between these three classes and do not preserve the geometry of the scene. This is underlined by the salt and pepper error presents in their land cover maps. Conversely, we can observe that~\method{} supplies a sharper (i.e., more homogeneous) characterization of the three land cover classes, especially concerning the forest class. 

The second example (Figures~\ref{fig:gard_ex2_b}, \ref{fig:gard_ex2_rf}, \ref{fig:gard_ex2_lstm} and~\ref{fig:gard_ex2_duplo}) represents a rural area principally characterized by different crop types. 
In this case we highlight a barley crop, in order to confirm the strong improvement in performance of \method{} upon $RF$ and $LSTM$ on this class, already emerged from the quantitative analysis (cf. Table~\ref{tab:PerClass_fm_gard}).
It is evident how both $RF$ and $LSTM$ fail to correctly classifying the \textit{Barley Crop}, mistakenly identifying it as \textit{Corn Crop} (interleaved by some \textit{Meadow}). Conversely, it can be seen how \method{} correctly identifies the correct extent of the \textit{Barley Crop}.

The third example (Figures~\ref{fig:gard_ex3_b}, \ref{fig:gard_ex3_rf}, \ref{fig:gard_ex3_lstm} and~\ref{fig:gard_ex3_duplo}) involves an urban area, mixed with rural areas and forest.
We highlight a large asphalted street which diagonally cuts the selected area. It can be noted how both $RF$ and $LSTM$ classify the street as water while, \method{} correctly classifies it as \textit{Urban Areas}. We remind that in our nomenclature we do not have a land cover class for street and, for this reason, detect a street as \textit{Urban Areas} is a reasonable compromise in our scenario.
 
Summarizing, on this study area, the qualitative analysis of the land cover maps demonstrates the effectiveness of \method{} compared to the other approaches on some specific classes like \textit{Barley Crop} and \textit{Corn Crop}, confirming the quantitative results reported in Table~\ref{tab:PerClass_fm_gard}. The analysis also shows how \method{} provides sharper and spatially coherent classification in mixed areas with respect to competitors which provide land cover maps affected by evident salt and pepper errors.

As concerns the~\textit{Reunion Island} study site, the first example (Figures~\ref{fig:reunion_ex1_b}, \ref{fig:reunion_ex1_rf}, \ref{fig:reunion_ex1_lstm} and~\ref{fig:reunion_ex1_duplo}) depicts a forest area. 
It can be noted how both $RF$ and $LSTM$, similarly to what observed for the first \textit{Gard} example (Figures~\ref{fig:gard_ex1_b}--~\ref{fig:gard_ex1_duplo}), are not able to preserve the geometry of the scene, erroneously placing sugar cane and orchards areas among the forest ones. Conversely, \method{} confirms its ability to produce a sharper and homogeneous demarcation of the forest.

The second example (Figures~\ref{fig:reunion_ex2_b}, \ref{fig:reunion_ex2_rf}, \ref{fig:reunion_ex2_lstm} and~\ref{fig:reunion_ex2_duplo}) shows a coastal area, where we highlight two urban settlements. The salt and pepper error that has been found to characterize the land cover maps produced by $RF$ and $LSTM$ can be clearly observed, with spots of meadow, orchards, greenhouse crops and sugar cane in the middle of urban areas, while the land cover map produced by \method{} depicts far more clear urban areas.

In the third example (Figures~\ref{fig:reunion_ex3_b}, \ref{fig:reunion_ex3_rf}, \ref{fig:reunion_ex3_lstm} and~\ref{fig:reunion_ex3_duplo}) a bare rocks area can be observed, produced by a lava flow of the \textit{Piton de la Fournaise} volcano on the eastern side of the island. It can be seen how $RF$ and $LSTM$ place urban areas and water in the middle of the bare rocks while, \method{} correctly identifies all the rocky area. It is interesting to observe how \method{} correctly identifies the border between the ocean and the mainland, while both competitors fail.

To sum up, the qualitative inspection of the land cover maps produced for the \textit{Reunion Island} study site confirms the quantitative results discussed in Section~\ref{sec:comparison} and, consolidates the observations drawn when discussing the land cover maps produced for \textit{Gard}, especially for what concerns the ability of \method{} to produce  sharper and more spatially coherent land cover maps with respect to both competitors. 

In order to have a wider example of how the land cover classification produced by \method{} differs from the ones provided by competing methods,
the land cover maps produced by \method{}, $RF$ and $LSTM$ on the \textit{Reunion Island} study site can be explored on our website~\footnote{\url{https://193.48.189.134/index.php/view/map/?repository=duplo&project=duplo}\\ Be aware that predictions are not available on some areas of the map (i.e. on the right side of the volcano) where there were not enough data available to perform the temporal gapfilling preprocessing on the SITS due to cloud issues (cf.~Section \ref{sec:data}).}.

\section{Conclusion}
\label{sec:conclu}
In this paper, a novel Deep Learning architecture to deal with optical Satellite Image Time Series classification has been proposed. The approach, named \method{}, leverages complementary representation of the remote sensing data to obtain a set of descriptors able to well discriminate the different land cover classes. The two-branch architecture involves a CNN and a RNN branch that process the same stream of information and, due to the difference in their structures, produces a more diverse and complete representation of the data. The final land cover classification is achieved by concatenating the features extracted by each branch. The framework is learned end-to-end from scratch.

The evaluation on two real-world study sites has shown that \method{} achieves better quantitative and qualitative results than state of the art classification methods for optical SITS data. In addition, the visual inspection of the land cover maps has advocated the effectiveness of our strategy.

As future work, we plan to extend the proposed approach towards the integration of other type of remote sensing data considering a multi-source scenario. For instance, our Deep Learning strategy can be extended to combine optical and radar SITS (i.e. Sentinel-2 and Sentinel-1 data) for land cover classification.

\section{Acknowledgements}
This work was supported by the French National Research Agency under the Investments for the Future Program, referred as ANR-16-CONV-0004 (DigitAg), the GEOSUD project with reference ANR-10-EQPX-20, the Programme National de T\'{e}l\'{e}d\'{e}tection Spatiale (PNTS, \url{http://www.insu.cnrs.fr/pnts} ), grant $n^{o}$PNTS-2018-5, as well as from the financial contribution from the French Ministry of agriculture ``Agricultural and Rural Development" trust account.

\bibliographystyle{elsarticle-num}
\bibliography{main}

\end{document}